\documentclass[final,1p,times]{elsarticle}

\usepackage{amssymb}
\usepackage{amsmath}
\usepackage{multirow}
\usepackage{subfig}
\usepackage{acronym}

\usepackage{pgfplots}
\pgfplotsset{compat=1.17}
\usepackage{tikz}
\usepackage{lscape}
\usepackage{tabularray}
\UseTblrLibrary{booktabs}
\usepackage{float}
\usepackage{makecell}
\usepackage{pgfplotstable}
\usepgfplotslibrary{fillbetween}

\usepackage{lineno}
\journal{Computers \& Chemical Engineering Journal}
\begin{document}

\begin{frontmatter}

\title{Predicting Filter Medium Performances in Chamber Filter Presses with Digital Twins Using Neural Network Technologies} %

\author[1,2]{Dennis Teutscher}
\author[3]{Tyll Weber-Carstanjen}
\author[1,4]{Stephan Simonis}
\author[1,2,4]{Mathias J.\ Krause}
\affiliation[1]{organization={Lattice Boltzmann Research Group, Karlsruhe Institute of Technology},%
    addressline={Englerstr.~2}, 
    city={Karlsruhe},
    postcode={76131}, 
    state={Baden-Württemberg},
    country={Germany}}
\affiliation[2]{organization={Institute for Mechanical Process Engineering and Mechanics, Karlsruhe Institute of Technology},%
    addressline={Straße am Forum~8}, 
    city={Karlsruhe},
    postcode={76131}, 
    state={Baden-Württemberg},
    country={Germany}}
\affiliation[4]{organization={Institute for Applied and Numerical Mathematics, Karlsruhe Institute of Technology},%
    addressline={Englerstr.~2}, 
    city={Karlsruhe},
    postcode={76131}, 
    state={Baden-Württemberg},
    country={Germany}}
\affiliation[3]{organization={Simex Filterpressen GmbH \& Co.\ KG},%
    addressline={Leibnizstr.~1}, 
    city={Calw},
    postcode={75365}, 
    state={Baden-Württemberg},
    country={Germany}}

\begin{abstract}
Efficient solid-liquid separation is crucial in industries like mining, but traditional chamber filter presses depend heavily on manual monitoring, leading to inefficiencies, downtime, and resource wastage. This paper introduces a machine learning-powered digital twin framework to improve operational flexibility and predictive control. A key challenge addressed is the degradation of the filter medium due to repeated cycles and clogging, which reduces filtration efficiency. To solve this, a neural network-based predictive model was developed to forecast operational parameters, such as pressure and flow rates, under various conditions. This predictive capability allows for optimized filtration cycles, reduced downtime, and improved process efficiency. Additionally, the model predicts the filter medium’s lifespan, aiding in maintenance planning and resource sustainability.
The digital twin framework enables seamless data exchange between filter press sensors and the predictive model, ensuring continuous updates to the training data and enhancing accuracy over time. Two neural network architectures, feedforward and recurrent, were evaluated. The recurrent neural network outperformed the feedforward model, demonstrating superior generalization. It achieved a relative $L^2$-norm error of $5\%$ for pressure and $9.3\%$ for flow rate prediction on partially known data. For completely unknown data, the relative errors were $18.4\%$ and $15.4\%$, respectively. Qualitative analysis showed strong alignment between predicted and measured data, with deviations within a confidence band of $8.2\%$ for pressure and $4.8\%$ for flow rate predictions. This work contributes an accurate predictive model, a new approach to predicting filter medium cycle impacts, and a real-time interface for model updates, ensuring adaptability to changing operational conditions.
\end{abstract}

\begin{graphicalabstract}
\includegraphics[width=1\linewidth]{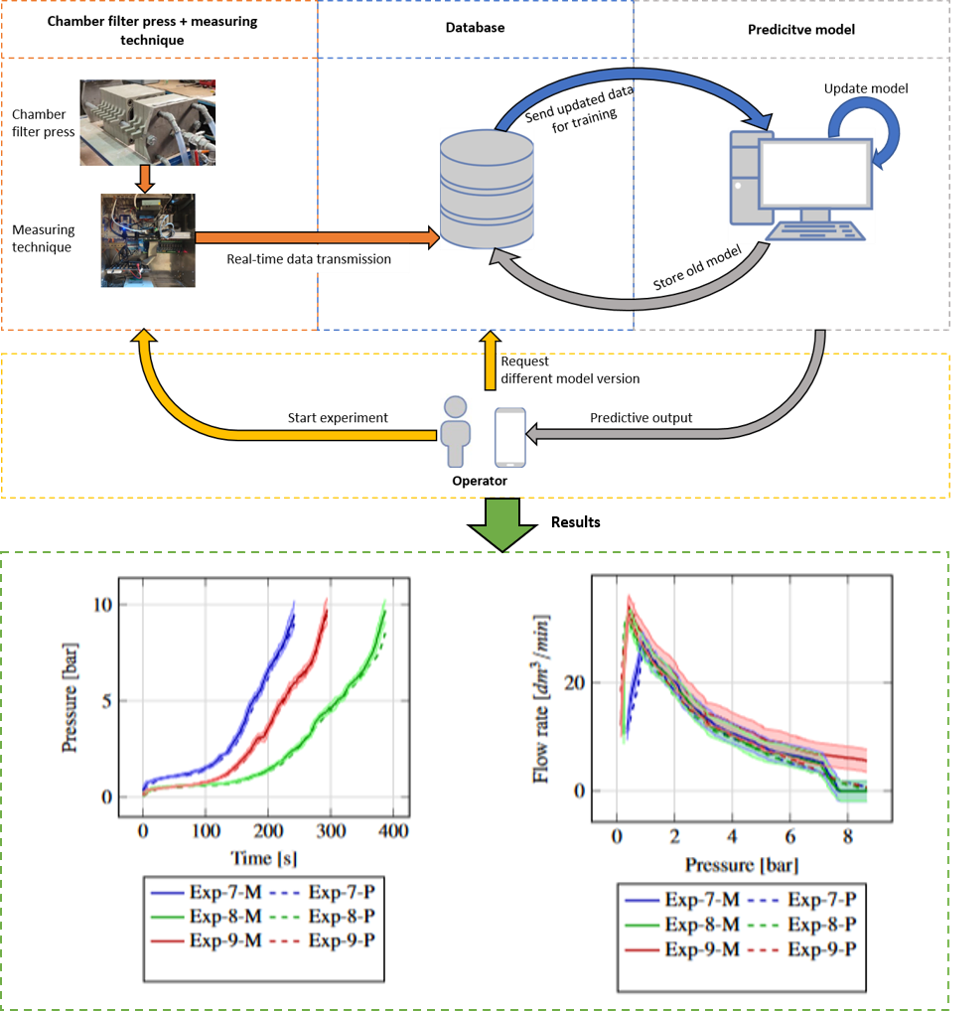}
\end{graphicalabstract}

\begin{highlights}
\item Machine learning-based digital twin framework of a chamber filter press for predicting key parameters like pressure and flow rates for different experiment configurations.
\item Predictive modeling estimates efficiency of filter medium, enabling proactive maintenance and resource sustainability.
\item Achieved robust predictive results, with relative $L^2$-norm errors of $5\%$ and $9.3\%$ for pressure and flow rate predictions using training data, and $18.4\%$ and $15.4\%$ for predictions on unknown validation data.
\item Seamless data exchange and continuous model updates ensure adaptability to evolving operational conditions.
\end{highlights}


\end{frontmatter}

\section{Introduction}

Filter presses play a critical role in numerous industries, including wastewater treatment, pharmaceuticals, and mining, by efficiently separating solids from liquids. In the mining sector, where vast quantities of ore are processed daily and water consumption is significant~\cite{GARNER12}, filter presses are indispensable for reducing fluid content in the separated material~\cite{GUNSON12}. This reduction not only minimizes water usage but also mitigates the risks associated with sludge storage, such as dam failures~\cite{BAKER17}.
As mining operations can have a big impact on the enviroment~\cite{MATSCHULLAT12}, it faces increasing environmental regulations and demands for sustainability. The optimization of filtration processes through innovations such as \ac{AI} and \ac{ML}, presents an opportunity to improve the automation, monitoring, as well as optimization of these filtration processes~\cite{TRAN21, MCCOY19}. By embedding predictive capabilities within filter press systems, operators can achieve higher precision in process control, leading to greater efficiency and better quality outputs.
This idea has been researched through different approaches. For example, Landman~\textit{et al.}~\cite{LANDAMAN97} used the theory of compressive rheology to predict the filtration time and maximize suspension throughput. Other approaches utilize \ac{NN} to predict metrics such as filtered volume~\cite{PUIGBARGUES12}, flow rate, and turbidity~\cite{HAWARI16}, or to predict and optimize particle counts~\cite{GRIFFITHS11}.
Another possibility is the application of \ac{CFD}. For instance, Spielman~\textit{et al.}~\cite{SPIELMAN68} employed a numerical model based on the Brinkman equation to predict pressure drops and filtration efficiency. Highly performant simulation software based on the \ac{LBM} could also be utilized for this purpose~\cite{OPENLB21}. However, the setup required for such simulations can be very challenging, covering aspects such as sedimentation~\cite{KRAUSE17,TRUNK21}, behavior during the filling phase, and actual filtering processes aided by porous areas~\cite{HAFEN22}. Additionally, while \ac{LBM} are significantly faster than other simulation approaches~\cite{HAUSSMANN20}, the results are still not available in real-time.  

Traditional filter press operations are highly dependent on manual monitoring and adjustments, which often result in inefficiencies, human errors, and increased downtime. Predicting performance parameters such as pressure and flow rates in real time is challenging, as these variables are influenced by fluctuating operating conditions and the state of the filter medium. The condition of the filter medium, particularly its level of clogging and the number of operational cycles it has undergone, significantly affects performance~\cite{CALLE2001,KANDRA2014518,kehat1967clogging, FRANKLE21,CALLE200240}.
Filter press dynamics are inherently complex due to the interplay of multiple nonlinear variables, including pressure, flow rate, and cake resistance, which evolve over time. In addition, fluctuations in components, such as membrane pumps, introduce noise, further complicating real-time predictive analysis. Existing data acquisition systems often lack the ability to seamlessly interface with advanced \ac{ML} algorithms, limiting operational flexibility and optimization potential. 
According to the work of McCoy~\textit{et al.}~\cite{MCCOY19} one problem in using \ac{ML} in the mining sector is the missing amount of data to train a model.
Addressing these challenges requires the integration of real-time data analytics with robust \ac{ML} capabilities to enable smarter and more adaptive process control.

The objective of this paper is to design, develop and validate an \ac{ML} model, integrated within a \ac{DT} framework, to accurately predict key operational parameters of a chamber filter press, specifically pressure and flow rate. The predictive model will enable proactive control and optimization of the filtration process, reducing the reliance on manual oversight, and ensuring consistent process quality. In addition, the \ac{DT} will focus on assessing the condition of the filter medium over repeated cycles, predicting its performance, and estimating the number of cycles it can sustain before replacement is required.

By addressing both real-time and historical data, the model will assist operators in determining optimal filter medium utilization, thereby enhancing efficiency and sustainability. Key performance metrics, including \ac{RMSE} and \ac{MSE}, are used to evaluate the accuracy of the model in training and validation datasets. 
In the remainder of the paper, first the methodology is described in Section~\ref{sec:meth}, which contains the experimental setup, parameter selection, and the architecture of the \ac{DT}, as well as the \ac{NN} model. Lastly, the results are discussed in Section~\ref{sec:res} and conclusions are drawn in Section~\ref{sec:conc}.

\section{Methodology}
\label{sec:meth}

\subsection{Experimental setup of the chamber filter press}
\begin{figure}[h]
\subfloat[]{
    \includegraphics[width=0.45\linewidth]{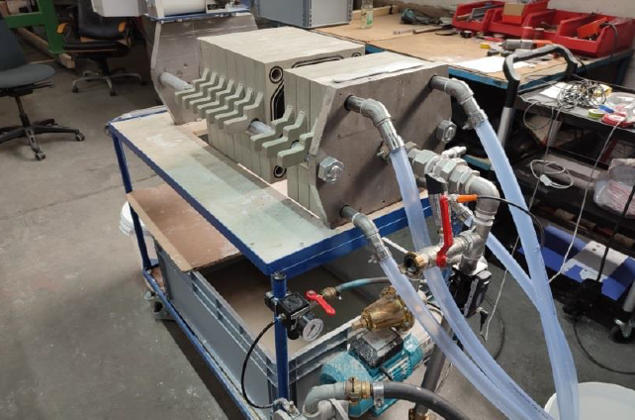}
}
\subfloat[]{
    \includegraphics[width=0.45\linewidth]{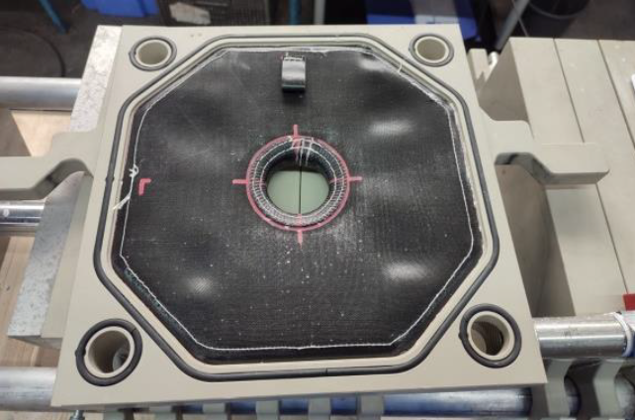}
}
\caption{The experimental setup of the chamber filter press, where (a) shows the filter press in its entirety and (b) shows a filter chamber with the filter cloth.}
\label{fig:experiment_setup}
\end{figure}

The chamber filter press used in this study has a plate size of $300~mm$, a compromise between typical small-scale test presses ($150~mm$ plate size) and larger versions (up to $1,000~mm$). This size is compatible with existing infrastructure, requires manageable suspension quantities, and is portable. The press is equipped with a membrane pump and a manually operated hydraulic cylinder, common configurations in industry, ensuring transferability of results. The setup includes a central inlet and outlets in each corner of the plates, maintains uniform conditions across the press, and aligns with industrial applications, such as automotive paint-sludge treatment and marine scrubber systems. This design ensures leak-tightness and consistent filtrate flow, enhancing reproducibility. Air blowing for filter cake removal is controlled through adjustable valves, allowing for variable air flow. Figure~\ref{fig:experiment_setup} shows the experimental setup next to the filter chamber with the filter cloth.

In order to ensure consistency and reproducible results that can be used to train a model, a suspension consisting of a mixture of water and perlite was used as the test material. 

The filter cloth used in the experiments had a throughput capacity of $5~\frac{l}{dm^2\cdot min}$. During each experiment, filtration was performed under controlled conditions to evaluate key parameters such as filtrate flow rate, pressure, filter cake formation, and overall filtration performance. The filtration process continued until a specified end point, such as a defined pressure or filtrate volume. Air blowing was then applied to remove the filter cake, with the air flow rate adjusted via the system’s valves.

\subsection{Parameter Selection}

Effective training of a \ac{NN} for a chamber filter press requires selecting input and output variables that capture the intricate dynamics of the filtration process. The flow rate of the filtrate and the operating pressure were chosen as the primary output variables because they are key indicators of the filtration performance. As filtration progresses, the pressure gradually increases, while the flow rate decreases, eventually approaching zero. This decrease in flow, combined with an increase in pressure, signals that the filter chambers are filled with accumulated solids, indicating the end of the filtration cycle. Accurately predicting these variables allows to monitor process efficiency in real time.

The input variables selected reflect various factors that influence the dynamics of filtration. The number of filter chambers directly impacts the filtration capacity, with a larger number of chambers enabling higher throughput. Filtration time captures the time-dependent evolution of flow and pressure, which is critical to understanding the progression of the cycle. The concentration of solids in the suspension plays a crucial role, as higher concentrations lead to a faster accumulation of solids within the chambers, affecting pressure dynamics and accelerating clogging. Another important input is the cycle count of the filter cloths, as repeated use degrades their performance, reducing the filtration efficiency over time. In addition, the maximum operating pressure sets the upper limit of the system, influencing the pressure and flow profiles throughout the process.

To enhance the generalization capability of the model, we constructed a training dataset to cover a wide range of operational conditions, including variations in chamber configurations, solid concentrations, and filter cloth usage. This comprehensive approach ensures that the model can accurately predict filtration outcomes across diverse scenarios, supporting effective, data-driven process control and optimization.

\subsection{Digital twin architecture}
\begin{figure}[ht]
    \centering
    \includegraphics[width=1.1\linewidth]{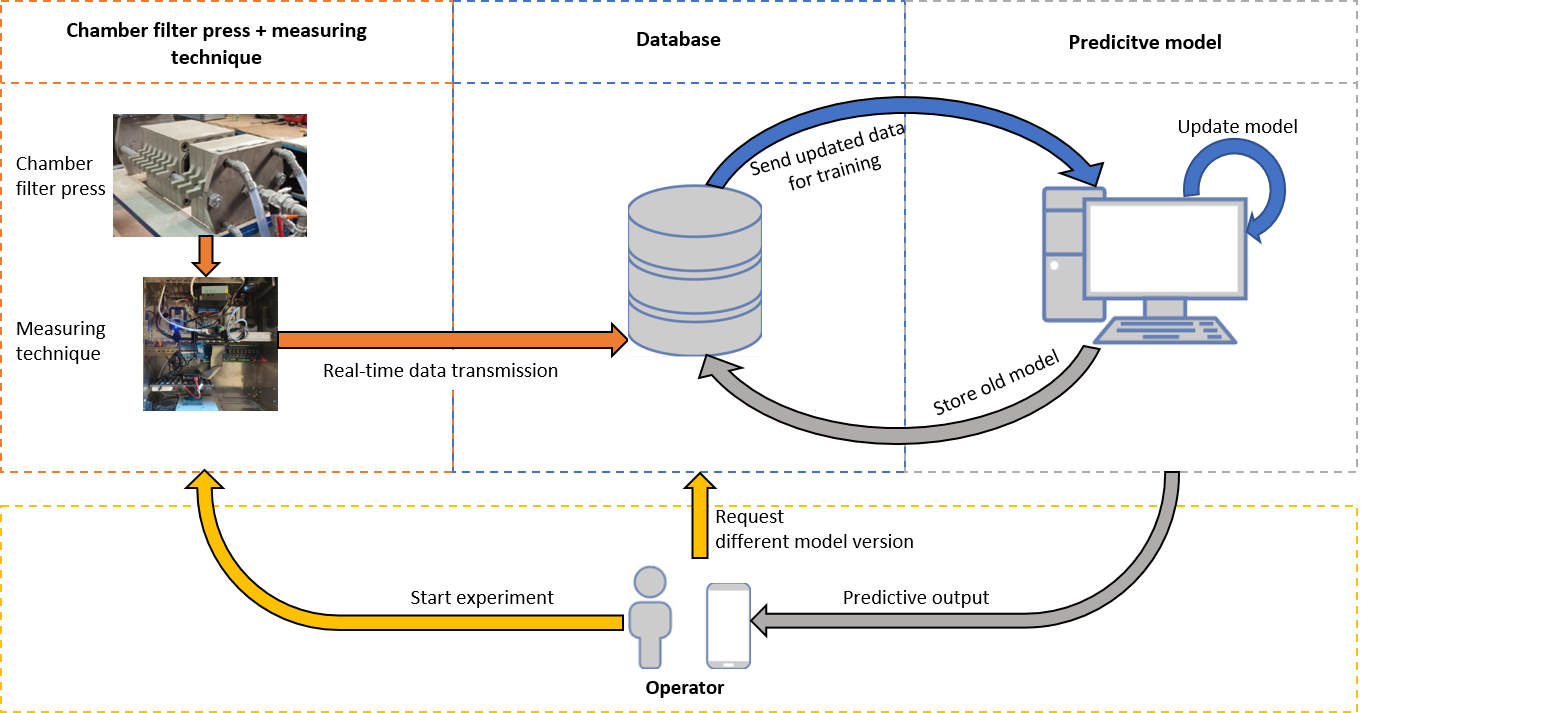}
    \caption{Architecture of the \ac{DT} framework for a chamber filter press, illustrating the communication flow between the filter press, real-time measuring techniques, the central database, the predictive model, and the operator.}
    \label{fig:dt_concept}
\end{figure}

The concept of the \ac{DT} for the chamber filter press revolves around a continuously improving \ac{NN} model that is updated with new training data as the filter press operates. This dynamic updating process ensures that the model predictions become increasingly reliable and accurate over time.
As illustrated in Figure~\ref{fig:dt_concept}, the operator initiates the process by setting up a new experiment with the static parameters identified, including the number of filter chambers, the filter cloth cycle, the maximum operating pressure, and the suspension concentration. These parameters are sent simultaneously to the database and the \ac{NN} model. The model uses this input to predict the expected course of pressure and flow rate over time, providing the operator with an estimated filtration time and an efficiency forecast for the process. This predictive insight helps in planning and optimizing the filtration cycle.
During the filtration process, sensors installed on the filter press continuously monitor and record real-time data, specifically flow rate and pressure. This data is transmitted to the database and linked to the corresponding experiment entry, ensuring traceability and coherence between static parameters and dynamic measurements.
Once the filtration cycle is complete, the recorded data can be fed back into the \ac{NN} as new training data. This iterative feedback loop enhances the model’s accuracy by refining its ability to predict future filtration performance based on historical patterns and real-time observations. 
This continuous learning approach not only optimizes the performance of the filter press, but also supports proactive decision-making, reducing downtime and improving process outcomes through more accurate predictions and insights.

\begin{figure}[ht]
    \centering
    \includegraphics[width=0.7\linewidth]{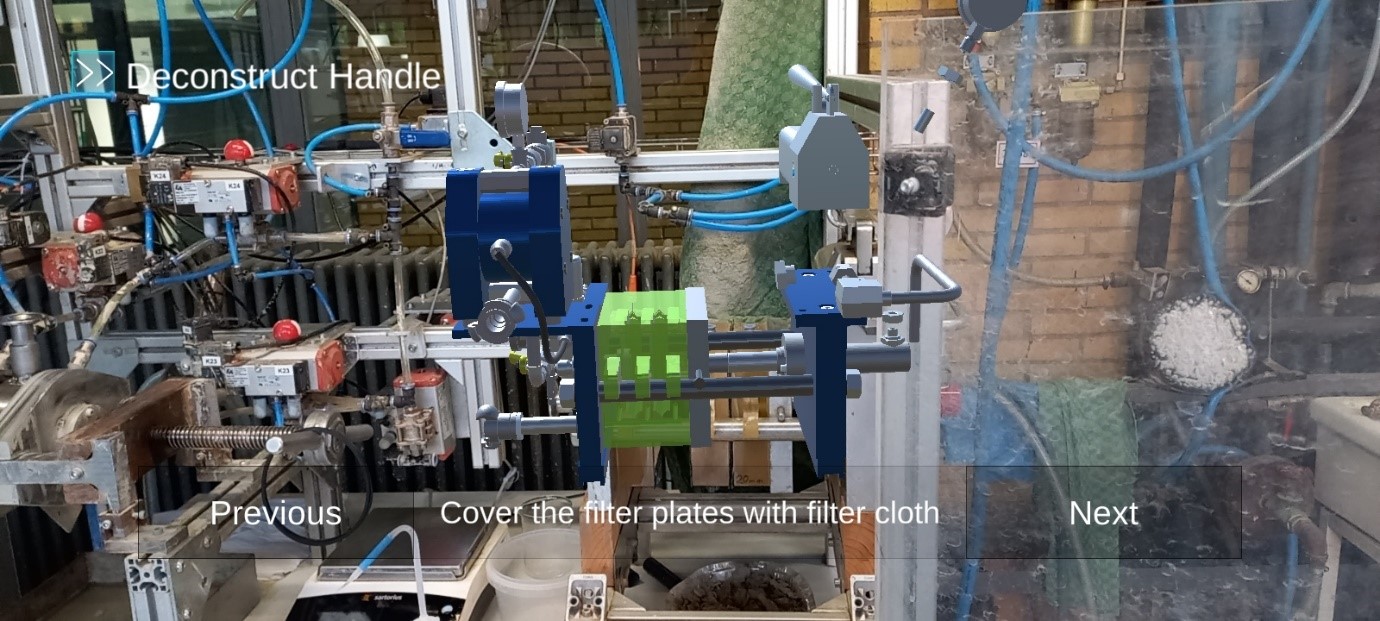}
    \caption{Integration of \ac{AR} with the \ac{DT} framework~\cite{TEUTSCHER22}. A three-dimensional model of the chamber filter press is overlaid on the real geometry, demonstrating the potential for visualizing operational states and diagnosing issues such as filter cloth wear or system errors in real time.}
    \label{fig:dt_AR}
\end{figure}

This \ac{DT} concept can also be expanded by integrating \ac{AR} technology. \ac{AR} could enable operators and maintenance personnel to visualize the state of the filter press directly on its physical counterpart, providing a real-time overlay of critical data such as performance metrics, sensor readings, or potential error states. For example, \ac{AR} could highlight issues such as wear or misalignment of the filter cloth or deviations in pressure distribution, allowing immediate troubleshooting. Furthermore, \ac{AR} could dynamically visualize the operational status of the filter press, including the progress of filtration cycles and system diagnostics, making complex data more intuitive and actionable.
Although this combination of \ac{DT} and \ac{AR} has significant potential to improve system understanding, troubleshooting, and maintenance, these aspects will not be explored further in this work. However, Figure~\ref{fig:dt_AR} illustrates a preliminary demonstration, from previous work, where a three-dimensional model of the chamber filter press is overlaid on its real geometry, showcasing the potential for future developments in this direction~\cite{TEUTSCHER22}.

\subsection{Experiments}

The experiments that serve as training and validation data for the \ac{NN} are summarized in Table~\ref{tab:trainingdata}, with a total of 34. The selection of these experiments was based on the need to cover a wide range of operating conditions and ensure that the \ac{NN} can learn patterns that generalize well to different scenarios. The configurations are designed to reflect a variety of process variables, such as concentration, filter plate number, end pressure, and filter medium cycles, which are expected to significantly influence the filtration process. By varying these parameters, we ensure that the network is exposed to a comprehensive set of inputs and can develop a robust understanding of the system's behavior.

\begin{table}[h!]
\centering
\caption{Available training and validation data from experiments.}
\begin{tabular}{c c c c c}
\toprule
\makecell{\textbf{Concentration [g/l]}} & \makecell{\textbf{Filter plate} \\ \textbf{number}} & \makecell{\textbf{End pressure} \\ \textbf{[bar]}} & \textbf{Cycles} & \textbf{Frequency} \\
\midrule
6.25                         & 2                            & 2.0                         & 34              & 1                  \\ \hline
6.25                         & 2                            & 4.0                         & 32              & 1                  \\ \hline
6.25                         & 2                            & 5.0                         & 31              & 1                  \\ \hline
6.25                         & 2                            & 6.0                         & 30              & 1                  \\ \hline
6.25                         & 2                            & 8.0                         & 29               & 1                  \\ \hline
12.50                        & 1                            & 10                          & 4              & 1                 \\ \hline
12.50                        & 2                            & 10.0                        & 2,4,5,6,7,14,23,35,36  & 9                  \\ \hline
12.50                        & 2                            & 7.0                         & 5              & 1                 \\ \hline
12.50                        & 2                            & 8.0                         & 6              & 1                  \\ \hline
12.50                        & 2                            & 0.2                         & 1              & 1                  \\ \hline
12.50                        & 2                            & 0.5                         & 10,11          & 2                  \\ \hline
12.50                        & 2                            & 0.7                         & 12,13          & 2                  \\ \hline
12.50                        & 3                            & 10                          & 1,2,3          & 3                  \\ \hline
25.00                        & 2                            & 10.0                        & 24             & 1                  \\ \hline
25.00                        & 2                            & 5.0                         & 18             & 1                  \\ \hline
25.00                        & 2                            & 6.0                         & 19             & 1                  \\ \hline
25.00                        & 2                            & 7.0                         & 20             & 1                  \\ \hline
25.00                        & 2                            & 8.0                         & 21             & 1                  \\ \hline
25.00                        & 2                            & 9.0                         & 22             & 1                  \\ \hline
25.00                        & 2                            & 10.0                        & 23             & 1                  \\ \hline
25.00                        & 3                            & 10.0                        & 25             & 1                  \\ \hline
25.00                        & 4                            & 10.0                        & 26             & 1                  \\ \hline
\bottomrule
\end{tabular}
\label{tab:trainingdata}
\end{table}

\subsection{Data logging and database development}

\begin{figure}[ht]
    \centering
    \includegraphics[width=1\linewidth]{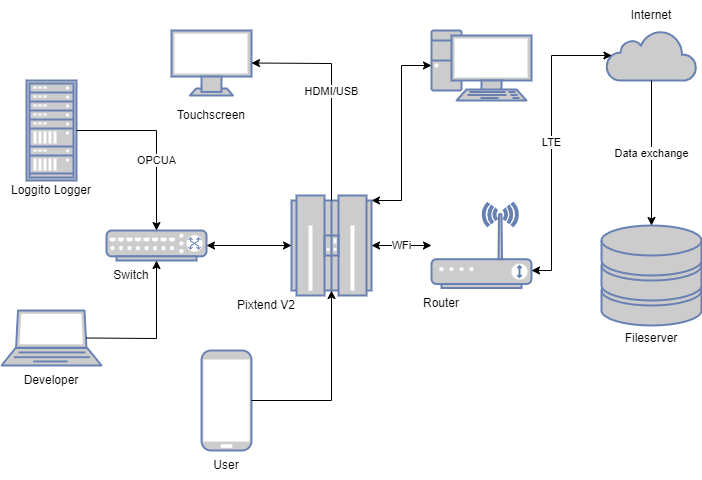}
    \caption{Schematic overview of the data acquisition and communication architecture for the chamber filter press system. The setup includes a Delphin data logger connected to sensors for pressure and flow rate measurement, which communicates with a control unit via OPC UA. The control system, represented by a PiXtend V2 board, interfaces with peripheral devices such as a touchscreen and user input terminals through HDMI/USB. Data transfer and remote monitoring are facilitated through a router, enabling Wi-Fi connectivity and LTE-based transmission to a fileserver for storage and further analysis. Developers and operators can access the system remotely or locally for control and data export.}
    \label{fig:concept_data_collection}
\end{figure}

The experimental filter press setup incorporates essential hardware components powered by a 24 V DC input, including a suspension pressure sensor, a flow sensor, and a delphin data logger. The data logger, which operates at a sampling rate of 10 Hz per channel with 24-bit resolution, is suitable for capturing high-resolution data necessary for the analysis of the filtration process. It features an integrated web server and supports the \ac{OPC UA} protocol, facilitating seamless data transmission and remote access (Figure~\ref{fig:concept_data_collection}). While the logger enables local data retrieval via USB using proprietary software, it does not inherently support the direct transmission of data to the model environment or provide control functionality for the active components of the filter press. Therefore, an independent control computer was introduced, equipped with a development board containing a quad-core 64-bit ARM Cortex A72 processor. This control system also includes an output module that interfaces with and manages the operation of the active elements in the filter press.

The control computer runs on a 64-bit Linux-based operating system and utilizes an InfluxDB time-series database for local data storage. Node-RED programming facilitates the management of local input and output (I/O) control and orchestrates the data exchange between the control computer and the data logger. Communication between the development board and the data logger is established over an Ethernet connection via the OPC UA protocol, where the data logger functions as the OPC UA server. Once data is transferred to the control system, it undergoes preprocessing, including filtering and analysis, to optimize the data storage load in the database.

The database architecture comprises two interconnected tables: one dedicated to experimental metadata and the other to measured data. The experiment table records user-defined parameters, including the experiment number, number of filter cycles, number of filter chambers, maximum operating pressure, and suspension concentration. These parameters are entered by the user via a graphical interface before initiating each experimental run. The measured data table contains time-stamped data points, such as pressure and flow rate readings, which are linked to the corresponding experiment via the experiment number as a foreign key. This relational structure allows for efficient organization and retrieval of experimental data.

Since InfluxDB utilizes time-based indexing, accurate synchronization between the control computer and the data logger is critical to maintaining data integrity. Initially, the \ac{NTP} is employed to synchronize the system clocks. However, given that NTP cannot achieve millisecond-level precision on the data logger, a \ac{PTP} server is subsequently initiated on the control computer. This ensures precise time synchronization between the two systems, which is essential for accurate data logging and analysis.

The system architecture is designed to accommodate various data transmission strategies, depending on the operational context of the filter press. For the prototype system described, the direct transmission of measurement data from the control system to the model cloud server via LTE was selected as the preferred method. This approach facilitates real-time monitoring and data exchange with external databases for further analysis and archiving.

\subsection{Neural network model}
\subsubsection{Selection of network type}

\begin{figure}[ht]
\centering
\subfloat[]{
        \begin{tikzpicture}[->, thick, every node/.style={circle, draw, minimum size=1cm}, node distance=1.5cm]

            \node[fill=black!20] (I1) at (0, 3) {};
            \node[fill=black!20] (I2) at (0, 1.5) {};
            \node[fill=black!20] (I3) at (0, 0) {};

            \node[fill=green!30] (H1) at (2, 3) {};
            \node[fill=green!30] (H2) at (2, 1.5) {};
            \node[fill=green!30] (H3) at (2, 0) {};

            \foreach \i in {1,2,3}
                \foreach \j in {1,2,3}
                    \draw[->] (I\i) -- (H\j);

            \draw[->] (H1) -- (4,3);
            \draw[->] (H2) -- (4,1.5);
            \draw[->] (H3) -- (4,0);

            \foreach \i in {1,2,3}
                \draw[->,looseness=4] (H\i) to [out=0,in=90] (H\i);
        \end{tikzpicture}
    }
\hspace{3cm}
    \subfloat[]{
        \begin{tikzpicture}[->, thick, every node/.style={circle, draw, minimum size=1cm}, node distance=1.5cm]

            \node[fill=black!20] (I1) at (0, 3) {};
            \node[fill=black!20] (I2) at (0, 1.5) {};
            \node[fill=black!20] (I3) at (0, 0) {};

            \node[fill=green!30] (H1) at (2, 3) {};
            \node[fill=green!30] (H2) at (2, 1.5) {};
            \node[fill=green!30] (H3) at (2, 0) {};

            \draw[->] (H1) -- (4,3);
            \draw[->] (H2) -- (4,1.5);
            \draw[->] (H3) -- (4,0);

            \foreach \i in {1,2,3}
                \foreach \j in {1,2,3}
                    \draw[->] (I\i) -- (H\j);
        \end{tikzpicture}
    }
    \caption{Comparison of \ac{RNN} (a) and \ac{FFNN} (b).
Gray nodes represent input neurons, and green nodes represent hidden neurons in both architectures.
In (a), the \ac{RNN} includes feedback loops, represented by the arrows looping back from the hidden neurons to themselves, enabling the processing of sequential data and temporal dependencies. The arrows pointing to the right indicate information flow to the output.
In (b), the \ac{FFNN} consists of direct connections between neurons, without feedback loops, illustrating a simpler network structure designed for static data processing.}
    \label{fig:network-comparison}
\end{figure}
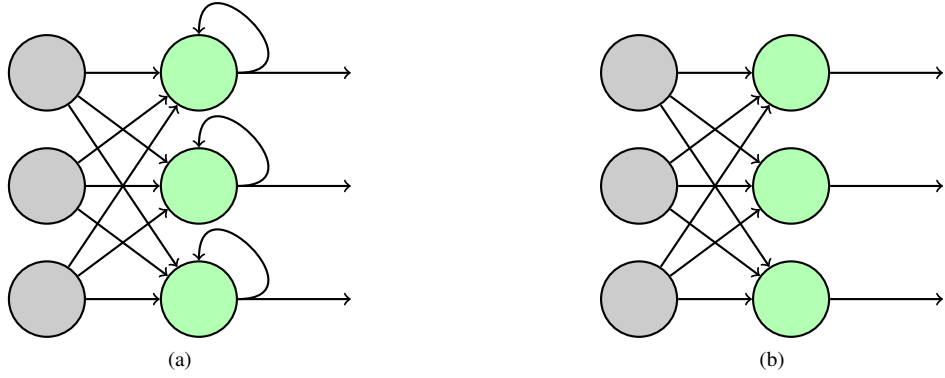

For the model, two \ac{NN} architectures were considered, the \ac{RNN}  and the \ac{FFNN}, the architecture of which is shown in Figure~\ref{fig:network-comparison}. \ac{FFNN}s are simple and effective for static data~\cite{SANDBERG01}, however, they lack the ability to model temporal dependencies and treat each input independently. \ac{RNN}s has the inherent ability to process sequential data, which could be essential to accurately modeling the dynamic behavior. \ac{RNN}s maintain an internal state that allows them to retain information about previous inputs~\cite{SHERSTINKSY20}, making them in theory well-suited for time-series data such as pressure and flow rate measurements. Its downside is that with enough accumulated data, it needs significantly more computational power to train. In this work both the \ac{FFNN} and \ac{RNN} were implemented in order to compare them against each other.

\subsubsection{Model architectures} 

The \ac{FFNN} consists of an input layer that accepts a normalized feature vector with five input variables. The model architecture includes two fully connected hidden layers with 64 and 32 neurons, respectively. Each hidden layer utilizes the rectified linear unit (ReLU) activation function to introduce nonlinearity and enhance the model's capability to learn complex relationships. The output layer comprises a single neuron with a linear activation function, suitable for predicting continuous target variables. The \ac{FFNN} is optimized using the Adam optimizer with a learning rate of $0.001$.

The \ac{RNN} is configured using a \ac{LSTM}~\cite{MIENYE24} layer to manage sequential data. The input layer receives sequences of length $10$, each consisting of five features per time step. The LSTM layer contains $64$ hidden units, enabling the model to capture temporal dependencies effectively. Following the LSTM layer, a fully connected output layer with a single neuron provides the final prediction, using a linear activation function for regression. Similar to the \ac{FFNN}, the \ac{RNN} is trained using the Adam optimizer with a learning rate of $0.001$. To facilitate temporal learning, the input data is prepared by generating sequences, allowing the network to leverage internal states for improved prediction accuracy. Data normalization and sequence preparation are essential preprocessing steps that enhance the performance of both models.

\subsubsection{Data preparation}
Before training the \ac{FFNN} or \ac{RNN}, the input data must undergo preprocessing~\cite{MCLARREN21}. The inputs are normalized for both networks using the following standardization formula:

\begin{equation} 
x_{\text{scaled}} = \frac{x - \mu}{\sigma}, 
\end{equation}
where $\mu$ represents the mean of the input variables and $\sigma$ denotes their standard deviation.

Although this normalization step is sufficient for training the \ac{FFNN}, the \ac{RNN} requires an additional preparation step: sequencing the input variables. This step structures the data into temporal sequences, enabling the \ac{RNN} to capture patterns over time and develop an internal representation of the expected behavior of the experiments. This sequencing enhances the RNNs ability to model temporal dependencies and improves predictive performance.

\subsubsection{Model evaluation}

The performance of the models was analyzed using key metrics such as the \ac{MSE}, the \ac{MAE}, and the coefficient of determination $R^2$. These metrics provide complementary insights into the performance of the model. \ac{MSE} quantifies the average squared differences between true and predicted values, penalizing larger errors more heavily than smaller ones. \ac{MAE} measures the average absolute difference between true and predicted values, treating all deviations equally, providing a more intuitive interpretation of the average error magnitude compared to \ac{MSE}. The coefficient of determination $R^2$ explains the proportion of variance in the true values captured by the model. The mathematical formulations of these error metrics are provided in Equations~\ref{eq:error_eval1}-\ref{eq:error_eval3}:

\begin{align}
    \text{MSE} &= \frac{1}{n} \sum_{i=1}^{n} \left( y_i - \hat{y}_i \right)^2, 
    \label{eq:error_eval1} \\
    \text{MAE} &= \frac{1}{n} \sum_{i=1}^{n} \left| y_i - \hat{y}_i \right|, 
    \label{eq:error_eval2} \\
    R^2 &= 1 - \frac{\sum_{i=1}^{n} \left( y_i - \hat{y}_i \right)^2}{\sum_{i=1}^{n} \left( y_i - \bar{y} \right)^2},
    \label{eq:error_eval3}
\end{align}
where $n$ is the total number of data points, $y_i$ represents the true values, and $\hat{y}_i$ denotes the predicted values. For $R^2$, $\bar{y}$ is the mean of the true values.

These metrics were tracked over the training epochs to evaluate the convergence of the model and to detect potential issues such as overfitting, where the model performs well on training data but poorly on validation data, or underfitting, where the model fails to capture the underlying data patterns.

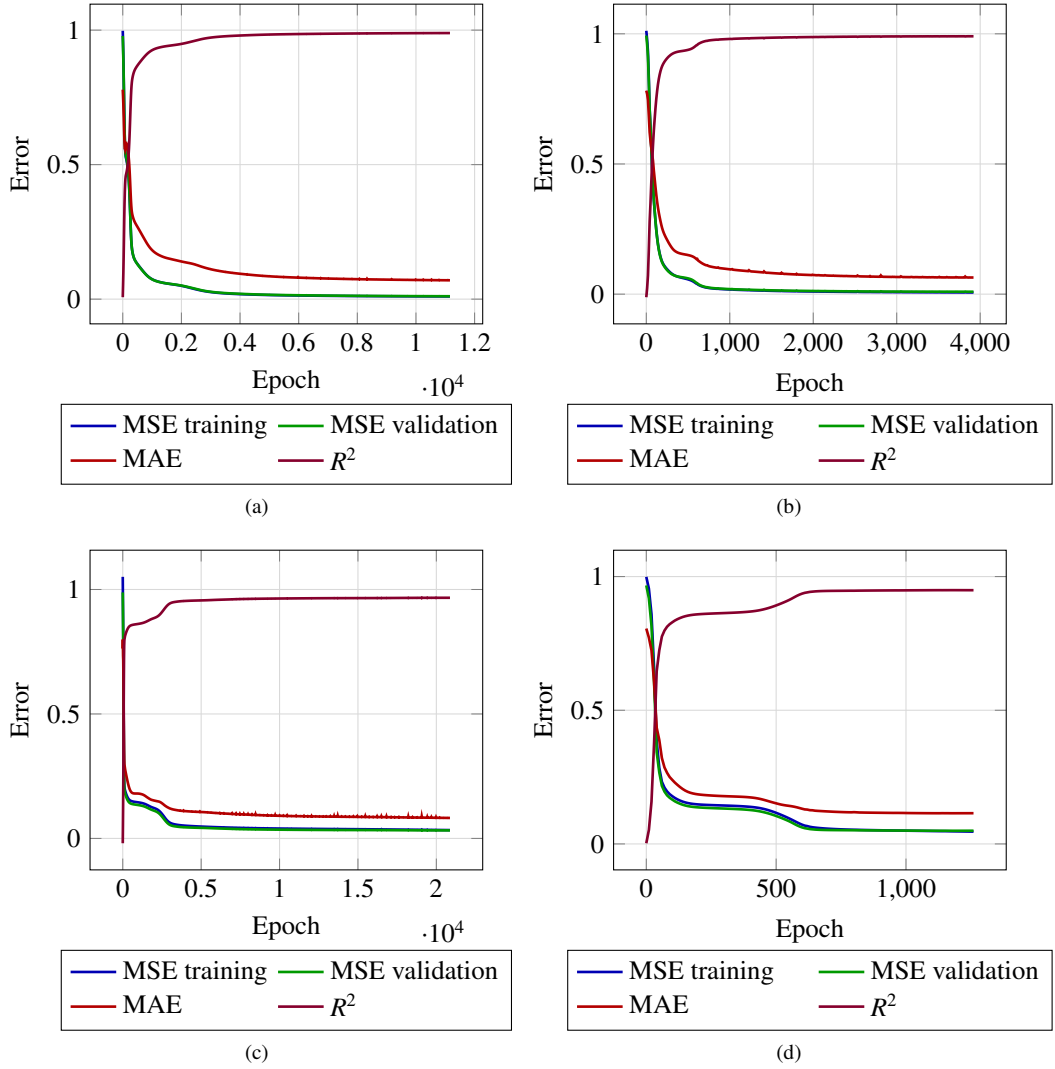
\begin{figure}[h!]
\subfloat[]{
\begin{tikzpicture}
\begin{axis}[
    width=0.5\textwidth, %
    xlabel={Epoch},
    ylabel={Error},
    legend style={
        at={(0.5,-0.25)}, %
        anchor=north,    
        legend columns=2, %
        /tikz/every even column/.append style={column sep=0.05cm} %
    },
    legend cell align={left}, %
    grid=both, %
    minor grid style={dotted}, %
    major grid style={solid, gray!30} %
]

\addplot[color=blue!70!black, line width=1pt] 
    table [x={epoch}, y={MSE_train}, col sep=comma] 
    {Pressure_MSE_MAE.csv};
\addlegendentry{MSE training}

\addplot[color=green!60!black, line width=1pt] 
    table [x={epoch}, y={MSE_test}, col sep=comma] 
    {Pressure_MSE_MAE.csv};
\addlegendentry{MSE validation}

\addplot[color=red!70!black, line width=1pt] 
    table [x={epoch}, y={MAE}, col sep=comma] 
    {Pressure_MSE_MAE.csv};
\addlegendentry{MAE}

\addplot[color=purple!70!black, line width=1pt] 
    table [x={epoch}, y={r2}, col sep=comma] 
    {Pressure_r2_score.csv};
\addlegendentry{$R^2$}

\end{axis}
\end{tikzpicture}
}
\hfill
\subfloat[]{
\begin{tikzpicture}
\begin{axis}[
    width=0.5\textwidth, %
    xlabel={Epoch},
    ylabel={Error},
    legend style={
        at={(0.5,-0.25)}, %
        anchor=north,    
        legend columns=2, %
        /tikz/every even column/.append style={column sep=0.5cm} %
    },
    legend cell align={left}, %
    grid=both, %
    minor grid style={dotted}, %
    major grid style={solid, gray!30} %
]
\addplot[color=blue!70!black, line width=1pt] 
    table [x={epoch}, y={MSE_train}, col sep=comma] 
    {Rec_Pressure_MSE_MAE.csv};
\addlegendentry{MSE training}

\addplot[color=green!60!black, line width=1pt] 
    table [x={epoch}, y={MSE_test}, col sep=comma] 
    {Rec_Pressure_MSE_MAE.csv};
\addlegendentry{MSE validation}

\addplot[color=red!70!black, line width=1pt] 
    table [x={epoch}, y={MAE}, col sep=comma] 
    {Rec_Pressure_MSE_MAE.csv};
\addlegendentry{MAE}

\addplot[color=purple!70!black, line width=1pt] 
    table [x={epoch}, y={r2}, col sep=comma] 
    {Rec_Pressure_r2_score.csv};
\addlegendentry{$R^2$}

\end{axis}
\end{tikzpicture}
}
\\
\subfloat[]{
\begin{tikzpicture}
\begin{axis}[
    width=0.5\textwidth, %
    xlabel={Epoch},
    ylabel={Error},
    legend style={
        at={(0.5,-0.25)}, %
        anchor=north,    
        legend columns=2, %
        /tikz/every even column/.append style={column sep=0.05cm} %
    },
    legend cell align={left}, %
    grid=both, %
    minor grid style={dotted}, %
    major grid style={solid, gray!30} %
]

\addplot[color=blue!70!black, line width=1pt] 
    table [x={epoch}, y={MSE_train}, col sep=comma] 
    {Flow_MSE_MAE.csv};
\addlegendentry{MSE training}

\addplot[color=green!60!black, line width=1pt] 
    table [x={epoch}, y={MSE_test}, col sep=comma] 
    {Flow_MSE_MAE.csv};
\addlegendentry{MSE validation}

\addplot[color=red!70!black, line width=1pt] 
    table [x={epoch}, y={MAE}, col sep=comma] 
    {Flow_MSE_MAE.csv};
\addlegendentry{MAE}

\addplot[color=purple!70!black, line width=1pt] 
    table [x={epoch}, y={r2}, col sep=comma] 
    {Flow_r2_score.csv};
\addlegendentry{$R^2$}

\end{axis}
\end{tikzpicture}
}
\hfill
\subfloat[]{
\begin{tikzpicture}
\begin{axis}[
    width=0.5\textwidth, %
    xlabel={Epoch},
    ylabel={Error},
    legend style={
        at={(0.5,-0.25)}, %
        anchor=north,    
        legend columns=2, %
        /tikz/every even column/.append style={column sep=0.5cm} %
    },
    legend cell align={left}, %
    grid=both, %
    minor grid style={dotted}, %
    major grid style={solid, gray!30} %
]
\addplot[color=blue!70!black, line width=1pt] 
    table [x={epoch}, y={MSE_train}, col sep=comma] 
    {Rec_Flow_MSE_MAE.csv};
\addlegendentry{MSE training}

\addplot[color=green!60!black, line width=1pt] 
    table [x={epoch}, y={MSE_test}, col sep=comma] 
    {Rec_Flow_MSE_MAE.csv};
\addlegendentry{MSE validation}

\addplot[color=red!70!black, line width=1pt] 
    table [x={epoch}, y={MAE}, col sep=comma] 
    {Rec_Flow_MSE_MAE.csv};
\addlegendentry{MAE}

\addplot[color=purple!70!black, line width=1pt] 
    table [x={epoch}, y={r2}, col sep=comma] 
    {Rec_Flow_r2_score.csv};
\addlegendentry{$R^2$}

\end{axis}
\end{tikzpicture}
}

\caption{Error metrics \ac{MSE}, MAE, and R2R2 for training and validation data during training of pressure and flow models. Figure (a) and (b) represent \ac{FFNN} and \ac{RNN} models for pressure prediction, respectively, while (c) and (d) depict \ac{FFNN} and \ac{RNN} models for flow prediction}
\label{fig:error}
\end{figure}

Figure~\ref{fig:error} illustrates the training and validation performance of the \ac{FFNN} and \ac{RNN} in predicting both the pressure and the flow rate. 
As shown in Figure~\ref{fig:error}(b), the \ac{RNN} model for pressure prediction demonstrates superior convergence and generalization compared to the \ac{FFNN} model (Figure~\ref{fig:error}(a)). The \ac{RNN} achieves training and validation MSEs of $0.0064$ and $0.0093$, respectively. Furthermore, \ac{MAE} stabilizes at $0.04$, and the $R^2$ score reaches $0.99$, indicating strong predictive performance. In contrast, \ac{FFNN} achieves a training \ac{MSE} of $0.01$ and a validation \ac{MSE} of $0.0108$, with a final $R^2$ score of $0.98$. Although the \ac{FFNN} demonstrates reasonable accuracy, the \ac{RNN} consistently achieves lower errors and faster convergence compared to the \ac{FFNN}.
For flow rate prediction, the \ac{FFNN} outperforms the \ac{RNN} in terms of error metrics, achieving training and validation MSEs of $0.031$ and $0.034$, respectively, along with an \ac{MAE} of $0.0823$ and an $R^2$ score of $0.967$. In contrast, the \ac{RNN} converges to training and validation MSEs of $0.049$ and $0.052$, with a final \ac{MAE} of $0.115$ and an $R^2$ score of $0.9485$. However, Figure~\ref{fig:error}(c) reveals that the validation \ac{MAE} of \ac{FFNN} exhibits spikes caused by fluctuations in the validation \ac{MSE}, which increases while the training \ac{MSE} decreases. This instability suggests overfitting and poorer generalization. In contrast, the \ac{RNN} model (Figure~\ref{fig:error}(d)) achieves a final test \ac{MSE} of $0.049$, closely aligned with the training \ac{MSE}, and an \ac{MAE} of $0.1156$. Unlike the \ac{FFNN}, the \ac{RNN} does not exhibit significant fluctuations in validation metrics, indicating better generalization and robustness.
Overall, the evaluation suggests that the \ac{RNN} model is better suited for this task, as it demonstrates superior generalization and stability across both prediction tasks.

\section{Results}
\label{sec:res}

\begin{figure}[h!]

\subfloat[]{
\begin{tikzpicture}
\begin{axis}[
    width=0.5\textwidth, %
    xlabel={Time [s]},
    ylabel={Pressure [bar]},
    legend style={
        at={(0.5,-0.25)}, %
        anchor=north,    
        legend columns=2, %
        /tikz/every even column/.append style={column sep=0.05cm} %
    },
    legend cell align={left}, %
    grid=both, %
    minor grid style={dotted}, %
    major grid style={solid, gray!30} %
]

\addplot[color=blue!70!black, line width=1pt] 
    table [x={Zeit (s)}, y={Druck (bar)}, col sep=comma] 
    {pressure_dat_16_02_RAW.csv};

\end{axis}
\end{tikzpicture}
}
\subfloat[]{
\begin{tikzpicture}
\begin{axis}[
    width=0.5\textwidth, %
    xlabel={Time [s]},
    ylabel={Pressure [bar]},
    legend style={
        at={(0.5,-0.25)}, %
        anchor=north,    
        legend columns=2, %
        /tikz/every even column/.append style={column sep=0.05cm} %
    },
    legend cell align={left}, %
    grid=both, %
    minor grid style={dotted}, %
    major grid style={solid, gray!30} %
]

\addplot[color=blue!70!black, line width=1pt] 
    table [x={Zeit (s)}, y={Druck (bar)}, col sep=comma] 
    {pressure_dat_23-01-16-02.csv};
\addplot[name path=1, color=blue!50!white, line width=0.5pt] 
    table [x={Zeit (s)}, y={upperBound}, col sep=comma]
    {pressure_dat_23-01-16-02.csv};

\addplot[name path=2, color=blue!50!white, line width=0.5pt] 
    table [x={Zeit (s)}, y={lowerBound}, col sep=comma]
    {pressure_dat_23-01-16-02.csv};

\addplot[blue!50!white, opacity=0.4] 
    fill between[of=1 and 2];

\end{axis}
\end{tikzpicture}
}
\caption{Depicted is the RAW data of a pressure trend in (a) and the MA pressure trend in (b) with CI90$\%$ bounds.}
\label{fig:raw_vs_MA}
\end{figure}
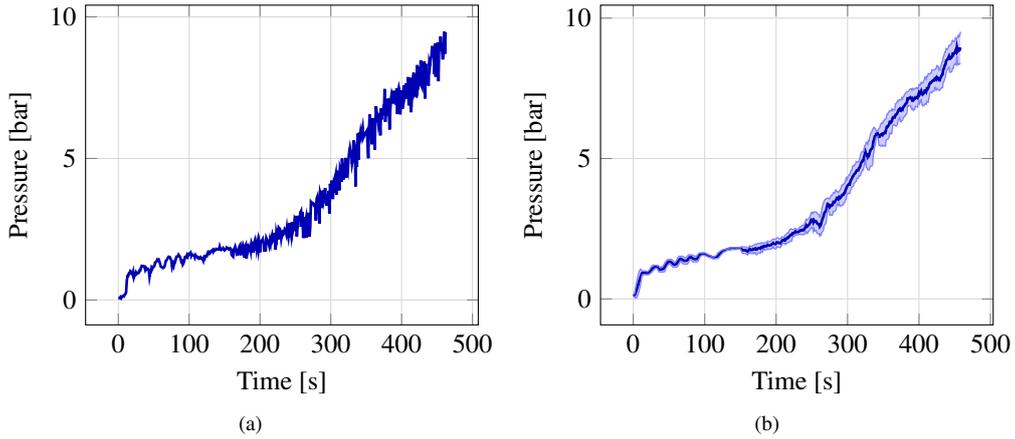

To evaluate the model, several key points must be addressed. In Figure~\ref{fig:raw_vs_MA}, the true pressure values of an experiment are shown, revealing significant fluctuations in the pressure trend (a). To fairly assess the prediction error, we first calculated a \ac{MA} and \ac{STD} for the experiments, as described by the following equations:

\begin{align}
    \text{MA}(t) &= \frac{1}{n} \sum_{i=t-n+1}^{t} x_i,
    \label{eq:MA} \\
    \text{STD} &= \sqrt{\frac{1}{n} \sum_{i=t-n+1}^t \left(x_i - \text{MA}(t)\right)^2},
    \label{eq:STD}
\end{align}
where \( n \) is the size of the averaging window, \( x_i \) are the data points, and \( n \) is the number of data points.

Next, a 90$\%$ confidence Interval (CI$90\%$) is defined using the calculated \ac{STD}:

\begin{equation}
    \text{CI90} = \text{MA} \pm z \cdot \text{STD},
\end{equation}

where \( z = 1.645 \) is the \(z\)-score corresponding to the 90\% percentile. This interval is used to assess how many predictions fall within the $CI90\%$.

By applying these equations to the experiments, an averaged line is obtained with upper and lower bounds, which represent the fluctuations in pressure and flow rate, as shown for the pressure in Figure~\ref{fig:raw_vs_MA} (b). This was done for the flow rate in the same manner.

In addition to \ac{MSE} and \ac{RMSE}, the \ac{RL2N} is also used to estimate the percentage error relative to the total magnitude of the experiments. It is defined as:

\begin{align}    
    \text{RMSE} &= \sqrt{\frac{1}{n} \sum_{i=1}^{n} (y_i - \hat{y}_i)^2},\\
    \text{RL2N} &= \frac{\sqrt{\sum_{i=1}^{n} (y_i - \hat{y}_i)^2}}{\sqrt{\sum_{i=1}^{n} y_i^2}}\cdot 100 ,
\end{align}
where \( \mathbf{y} \) represents the true values and \( \hat{\mathbf{y}} \) represents the predicted values.
In order to get an idea of the deviation of points in regards to the $CI90\%$ bounds the \ac{RL2N-B} is introduced and is defined as

\begin{align}
    \text{RL2N-B} &= \sqrt{\frac{\int e^2(x) \, dx}{\int f^2(x) \, dx}}\cdot 100,\\
    e^2(x) &= 
    \begin{cases} 
        0, & \text{if } \hat{f}(x) \in [l(x), u(x)], \\
        \min\big((\hat{f}(x) - l(x))^2, (\hat{f}(x) - u(x))^2\big), & \text{if } \hat{f}(x) \notin [l(x), u(x)].
    \end{cases}
\end{align}
where $e^2(x)$ represents the squared error function, which is zero if the predicted function $\hat{f}(x)$ lies within the confidence band $[l(x), u(x)]$ (lower bound, upper bound). Otherwise, $e^2(x)$ is computed as the square of the minimum distance between $\hat{f}(x)$ and the bounds of the confidence band. $\int f^2(x) , dx$ serves as the normalization factor, representing the total magnitude of the true function $f(x)$.

Last but not least the percentage of points inside the CI90$\%$ bounds (PIB) is also used for evaluation.

In this section, two experimental data sets are considered: Table~\ref{tab:experiments}, which contains experiments used for training and validation (split as 80\%/20\%), and Table~\ref{tab:experiments_val}, which contains completely unknown experiments, including one with an unknown concentration to the model.

\begin{table}[h!]
\centering
\caption{Experiment data with diverse configurations and filter cycles. The experiment are partially known to the model since it was split into $80\%$ training and $20\%$ validation data.}
\begin{tabular}{c c c c c}
\toprule
\textbf{Experiment} & \textbf{Concentration [g/l]} & \textbf{Filter plate number} & \textbf{End pressure (bar)} & \textbf{Cycles} \\
\midrule
1 & 12.5  & 2  &   10  & 2  \\ %
2 & 12.5  & 2  &   10  & 7  \\ %
3 & 12.5  & 2  &   10  & 35 \\ %
4 & 12.5  & 2  &   10  & 36\\ %
5 & 6.25  & 2  &    8  & 29 \\%'23-03-07-05' \\ \hline
6 & 25    & 2 &    10  &23 \\%23-03-06-05
7 & 12.5  & 1  &   10  & 4  \\ %
8 & 12.5  & 3  &   10  & 2  \\ %
9 & 12.5  & 2  &   10  & 5 \\%23-02-17-03
10& 25    & 1  &   10  &24\\%23-03-06-06
11& 25    & 3  &   10  &25\\%23-03-07-01
12& 25    & 4  &   10  &26\\%23-03-07-02
\bottomrule
\end{tabular}
\label{tab:experiments}
\end{table}

\begin{table}[h!]
\centering
\caption{Experiment data with diverse configurations and filter cycles. The experiments are completely unknown to the model.}
\begin{tabular}{c c c c c}
\toprule
\textbf{Experiment} & \textbf{Concentration [g/l]} & \textbf{Filter plate number} & \textbf{End pressure (bar)} & \textbf{Cycles} \\
\midrule
1 val & 6.25  & 2  &   10  & 24  \\ %
2 val & 12.5  & 2  &   10  & 30  \\ %
3 val & 12.5  & 2  &   10  & 11 \\ %
4 val & 12.5  & 2  &   10  & 10\\ %
5 val & 12.5  & 2  &   10  & 9  \\ %
6 val & 12.5  & 3  &   10  & 6  \\ %
7 val& 15    & 2  &   10  &7\\%23-09-08-002
8 val& 15    & 2  &   10  &8 \\ %
\bottomrule
\end{tabular}
\label{tab:experiments_val}
\end{table}

\subsubsection{Pressure prediction}

\paragraph{Partially known data}

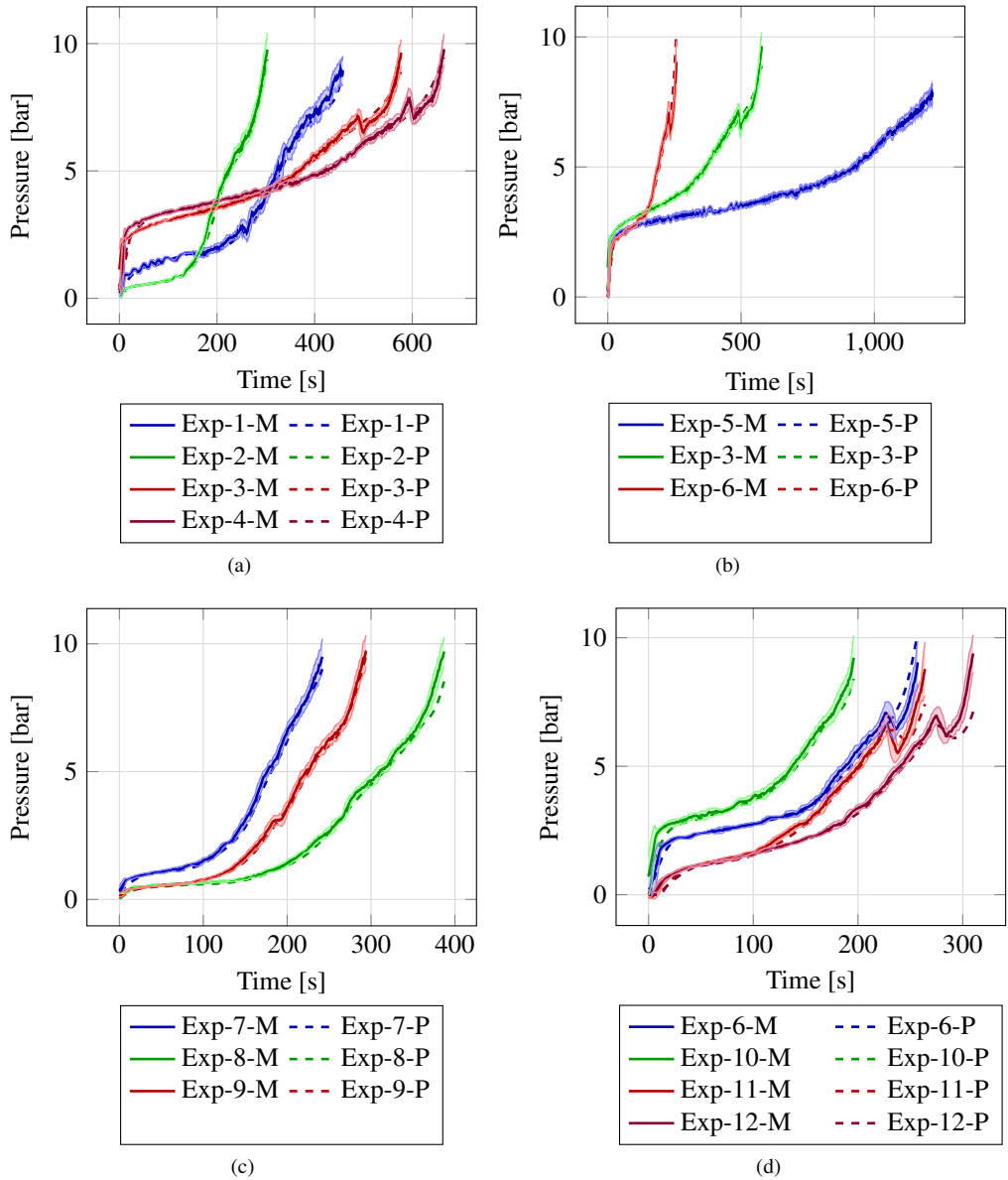
\begin{figure}[h!]

\subfloat[]{
\begin{tikzpicture}
\begin{axis}[
    width=0.5\textwidth, %
    xlabel={Time [s]},
    ylabel={Pressure [bar]},
    legend style={
        at={(0.5,-0.25)}, %
        anchor=north,    
        legend columns=2, %
        /tikz/every even column/.append style={column sep=0.05cm} %
    },
    legend cell align={left}, %
    grid=both, %
    minor grid style={dotted}, %
    major grid style={solid, gray!30} %
]

\addplot[color=blue!70!black, line width=1pt] 
    table [x={Zeit (s)}, y={Druck (bar)}, col sep=comma] 
    {pressure_dat_23-01-16-02.csv};
\addlegendentry{Exp-1-M}

\addplot[color=blue!70!black, line width=1pt, dashed] 
    table [x={Zeit (s)}, y={Pred (bar)}, col sep=comma]    
    {pressure_dat_23-01-16-02.csv};
\addlegendentry{Exp-1-P}

\addplot[color=green!60!black, line width=1pt] 
    table [x={Zeit (s)}, y={Druck (bar)}, col sep=comma] 
    {pressure_dat_23-02-20-01.csv};
\addlegendentry{Exp-2-M}

\addplot[color=green!60!black, line width=1pt, dashed] 
    table [x={Zeit (s)}, y={Pred (bar)}, col sep=comma]    
    {pressure_dat_23-02-20-01.csv};
\addlegendentry{Exp-2-P}

\addplot[color=red!70!black, line width=1pt] 
    table [x={Zeit (s)}, y={Druck (bar)}, col sep=comma] 
    {pressure_dat_23-03-14-01.csv};
\addlegendentry{Exp-3-M}

\addplot[color=red!70!black, line width=1pt, dashed] 
    table [x={Zeit (s)}, y={Pred (bar)}, col sep=comma]    
    {pressure_dat_23-03-14-01.csv};
\addlegendentry{Exp-3-P}

\addplot[color=purple!70!black, line width=1pt] 
    table [x={Zeit (s)}, y={Druck (bar)}, col sep=comma] 
    {pressure_dat_23-03-14-02.csv};
\addlegendentry{Exp-4-M}

\addplot[color=purple!70!black, line width=1pt, dashed] 
    table [x={Zeit (s)}, y={Pred (bar)}, col sep=comma]    
    {pressure_dat_23-03-14-02.csv};
\addlegendentry{Exp-4-P}

\addplot[name path=upper, color=blue!50!white, line width=0.5pt] 
    table [x={Zeit (s)}, y={upperBound}, col sep=comma]
    {pressure_dat_23-01-16-02.csv};

\addplot[name path=lower, color=blue!50!white, line width=0.5pt] 
    table [x={Zeit (s)}, y={lowerBound}, col sep=comma]
    {pressure_dat_23-01-16-02.csv};

\addplot[blue!50!white, opacity=0.4] 
    fill between[of=upper and lower];

\addplot[name path=upper2, color=green!50!white, line width=0.5pt] 
    table [x={Zeit (s)}, y={upperBound}, col sep=comma]
    {pressure_dat_23-02-20-01.csv};

\addplot[name path=lower2, color=green!50!white, line width=0.5pt] 
    table [x={Zeit (s)}, y={lowerBound}, col sep=comma]
    {pressure_dat_23-02-20-01.csv};

\addplot[green!50!white, opacity=0.4] 
    fill between[of=upper2 and lower2];

\addplot[name path=upper3, color=red!50!white, line width=0.5pt] 
    table [x={Zeit (s)}, y={upperBound}, col sep=comma]
    {pressure_dat_23-03-14-01.csv};

\addplot[name path=lower3, color=red!50!white, line width=0.5pt] 
    table [x={Zeit (s)}, y={lowerBound}, col sep=comma]
    {pressure_dat_23-03-14-01.csv};

\addplot[red!50!white, opacity=0.4] 
    fill between[of=upper3 and lower3];

\addplot[name path=upper4, color=purple!50!white, line width=0.5pt] 
    table [x={Zeit (s)}, y={upperBound}, col sep=comma]
    {pressure_dat_23-03-14-02.csv};

\addplot[name path=lower4, color=purple!50!white, line width=0.5pt] 
    table [x={Zeit (s)}, y={lowerBound}, col sep=comma]
    {pressure_dat_23-03-14-02.csv};

\addplot[purple!50!white, opacity=0.4] 
    fill between[of=upper4 and lower4];

\end{axis}
\end{tikzpicture}
}
\subfloat[]{
\begin{tikzpicture}
\begin{axis}[
    width=0.5\textwidth, %
    xlabel={Time [s]},
    ylabel={Pressure [bar]},
    legend style={
        at={(0.5,-0.25)}, %
        anchor=north,    
        legend columns=2, %
        /tikz/every even column/.append style={column sep=0.05cm} %
    },
    legend cell align={left}, %
    grid=both, %
    minor grid style={dotted}, %
    major grid style={solid, gray!30} %
]

\addplot[color=blue!70!black, line width=1pt] 
    table [x={Zeit (s)}, y={Druck (bar)}, col sep=comma] 
    {pressure_dat_23-03-07-05.csv};
\addlegendentry{Exp-5-M}

\addplot[color=blue!70!black, line width=1pt,dashed] 
    table [x={Zeit (s)}, y={Pred (bar)}, col sep=comma]    
    {pressure_dat_23-03-07-05.csv};
\addlegendentry{Exp-5-P}

\addplot[color=green!60!black, line width=1pt] 
    table [x={Zeit (s)}, y={Druck (bar)}, col sep=comma] 
    {pressure_dat_23-03-14-01.csv};
\addlegendentry{Exp-3-M}

\addplot[color=green!60!black, line width=1pt,dashed] 
    table [x={Zeit (s)}, y={Pred (bar)}, col sep=comma]    
    {pressure_dat_23-03-14-01.csv};
\addlegendentry{Exp-3-P}

\addplot[color=red!70!black, line width=1pt] 
    table [x={Zeit (s)}, y={Druck (bar)}, col sep=comma] 
    {pressure_dat_23-03-06-05.csv};
\addlegendentry{Exp-6-M}

\addplot[color=red!70!black, line width=1pt,dashed] 
    table [x={Zeit (s)}, y={Pred (bar)}, col sep=comma]    
    {pressure_dat_23-03-06-05.csv};
\addlegendentry{Exp-6-P}

\addplot[color=purple!70!black, line width=1pt, opacity=0] 
    table [x={Druck (bar)}, y={Druck (bar)}, col sep=comma]    
    {pressure_empty.csv}; %
\addlegendentry{\phantom{Exp}}

\addplot[name path=upper5, color=blue!50!white, line width=0.5pt] 
    table [x={Zeit (s)}, y={upperBound}, col sep=comma]
    {pressure_dat_23-03-07-05.csv};

\addplot[name path=lower5, color=blue!50!white, line width=0.5pt] 
    table [x={Zeit (s)}, y={lowerBound}, col sep=comma]
    {pressure_dat_23-03-07-05.csv};

\addplot[blue!50!white, opacity=0.4] 
    fill between[of=upper5 and lower5];

\addplot[name path=upper6, color=green!50!white, line width=0.5pt] 
    table [x={Zeit (s)}, y={upperBound}, col sep=comma]
    {pressure_dat_23-03-14-01.csv};

\addplot[name path=lower6, color=green!50!white, line width=0.5pt] 
    table [x={Zeit (s)}, y={lowerBound}, col sep=comma]
    {pressure_dat_23-03-14-01.csv};

\addplot[green!50!white, opacity=0.4] 
    fill between[of=upper6 and lower6];

\addplot[name path=upper7, color=red!50!white, line width=0.5pt] 
    table [x={Zeit (s)}, y={upperBound}, col sep=comma]
    {pressure_dat_23-03-06-05.csv};

\addplot[name path=lower7, color=red!50!white, line width=0.5pt] 
    table [x={Zeit (s)}, y={lowerBound}, col sep=comma]
    {pressure_dat_23-03-06-05.csv};

\addplot[red!50!white, opacity=0.4] 
    fill between[of=upper7 and lower7];

\end{axis}
\end{tikzpicture}
}

\subfloat[]{
\begin{tikzpicture}
\begin{axis}[
    width=0.5\textwidth, %
    xlabel={Time [s]},
    ylabel={Pressure [bar]},
    legend style={
        at={(0.5,-0.25)}, %
        anchor=north,    
        legend columns=2, %
        /tikz/every even column/.append style={column sep=0.05cm} %
    },
    legend cell align={left}, %
    grid=both, %
    minor grid style={dotted}, %
    major grid style={solid, gray!30} %
]

\addplot[color=blue!70!black, line width=1pt] 
    table [x={Zeit (s)}, y={Druck (bar)}, col sep=comma] 
    {pressure_dat_23-02-20-05.csv};
\addlegendentry{Exp-7-M}

\addplot[color=blue!70!black, line width=1pt, dashed] 
    table [x={Zeit (s)}, y={Pred (bar)}, col sep=comma]    
    {pressure_dat_23-02-20-05.csv};
\addlegendentry{Exp-7-P}

\addplot[color=green!60!black, line width=1pt] 
    table [x={Zeit (s)}, y={Druck (bar)}, col sep=comma] 
    {pressure_dat_23-02-20-04.csv};
\addlegendentry{Exp-8-M}

\addplot[color=green!60!black, line width=1pt,dashed] 
    table [x={Zeit (s)}, y={Pred (bar)}, col sep=comma]    
    {pressure_dat_23-02-20-04.csv};
\addlegendentry{Exp-8-P}

\addplot[color=red!70!black, line width=1pt] 
    table [x={Zeit (s)}, y={Druck (bar)}, col sep=comma] 
    {pressure_dat_23-02-17-03.csv};
\addlegendentry{Exp-9-M}

\addplot[color=red!70!black, line width=1pt,dashed] 
    table [x={Zeit (s)}, y={Pred (bar)}, col sep=comma]    
    {pressure_dat_23-02-17-03.csv};
\addlegendentry{Exp-9-P}

\addplot[color=purple!70!black, line width=1pt, opacity=0] 
    table [x={Zeit (s)}, y={Pred (bar)}, col sep=comma]    
    {pressure_empty.csv}; %

\addplot[name path=upper7, color=blue!50!white, line width=0.5pt] 
    table [x={Zeit (s)}, y={upperBound}, col sep=comma]
    {pressure_dat_23-02-20-05.csv};

\addplot[name path=lower7, color=blue!50!white, line width=0.5pt] 
    table [x={Zeit (s)}, y={lowerBound}, col sep=comma]
    {pressure_dat_23-02-20-05.csv};

\addplot[blue!50!white, opacity=0.4] 
    fill between[of=upper7 and lower7];

\addplot[name path=upper8, color=green!50!white, line width=0.5pt] 
    table [x={Zeit (s)}, y={upperBound}, col sep=comma]
    {pressure_dat_23-02-20-04.csv};

\addplot[name path=lower8, color=green!50!white, line width=0.5pt] 
    table [x={Zeit (s)}, y={lowerBound}, col sep=comma]
    {pressure_dat_23-02-20-04.csv};

\addplot[green!50!white, opacity=0.4] 
    fill between[of=upper8 and lower8];

\addplot[name path=upper9, color=red!50!white, line width=0.5pt] 
    table [x={Zeit (s)}, y={upperBound}, col sep=comma]
    {pressure_dat_23-02-17-03.csv};

\addplot[name path=lower9, color=red!50!white, line width=0.5pt] 
    table [x={Zeit (s)}, y={lowerBound}, col sep=comma]
    {pressure_dat_23-02-17-03.csv};

\addplot[red!50!white, opacity=0.4] 
    fill between[of=upper9 and lower9];

\addlegendentry{\phantom{Exp}}
\end{axis}
\end{tikzpicture}
}
\hfill
\subfloat[]{
\begin{tikzpicture}
\begin{axis}[
    width=0.5\textwidth, %
    xlabel={Time [s]},
    ylabel={Pressure [bar]},
    legend style={
        at={(0.5,-0.25)}, %
        anchor=north,    
        legend columns=2, %
        /tikz/every even column/.append style={column sep=0.5cm} %
    },
    legend cell align={left}, %
    grid=both, %
    minor grid style={dotted}, %
    major grid style={solid, gray!30} %
]

\addplot[color=blue!70!black, line width=1pt] 
    table [x={Zeit (s)}, y={Druck (bar)}, col sep=comma] 
    {pressure_dat_23-03-06-05.csv};
\addlegendentry{Exp-6-M}

\addplot[color=blue!70!black, line width=1pt, dashed] 
    table [x={Zeit (s)}, y={Pred (bar)}, col sep=comma]    
    {pressure_dat_23-03-06-05.csv};
\addlegendentry{Exp-6-P}

\addplot[color=green!60!black, line width=1pt] 
    table [x={Zeit (s)}, y={Druck (bar)}, col sep=comma] 
    {pressure_dat_23-03-06-06.csv};
\addlegendentry{Exp-10-M}

\addplot[color=green!60!black, line width=1pt,dashed] 
    table [x={Zeit (s)}, y={Pred (bar)}, col sep=comma]    
    {pressure_dat_23-03-06-06.csv};
\addlegendentry{Exp-10-P}

\addplot[color=red!70!black, line width=1pt] 
    table [x={Zeit (s)}, y={Druck (bar)}, col sep=comma] 
    {pressure_dat_23-03-07-01.csv};
\addlegendentry{Exp-11-M}

\addplot[color=red!70!black, line width=1pt,dashed] 
    table [x={Zeit (s)}, y={Pred (bar)}, col sep=comma]    
    {pressure_dat_23-03-07-01.csv};
\addlegendentry{Exp-11-P}

\addplot[color=purple!70!black, line width=1pt] 
    table [x={Zeit (s)}, y={Druck (bar)}, col sep=comma] 
    {pressure_dat_23-03-07-02.csv};
\addlegendentry{Exp-12-M}

\addplot[color=purple!70!black, line width=1pt, dashed] 
    table [x={Zeit (s)}, y={Pred (bar)}, col sep=comma]    
    {pressure_dat_23-03-07-02.csv};
\addlegendentry{Exp-12-P}

\addplot[name path=upper8, color=green!50!white, line width=0.5pt] 
    table [x={Zeit (s)}, y={upperBound}, col sep=comma]
    {pressure_dat_23-03-06-06.csv};

\addplot[name path=lower8, color=green!50!white, line width=0.5pt] 
    table [x={Zeit (s)}, y={lowerBound}, col sep=comma]
    {pressure_dat_23-03-06-06.csv};

\addplot[green!50!white, opacity=0.4] 
    fill between[of=upper8 and lower8];

\addplot[name path=upper15, color=red!50!white, line width=0.5pt] 
    table [x={Zeit (s)}, y={upperBound}, col sep=comma]
    {pressure_dat_23-03-07-01.csv};

\addplot[name path=lower15, color=red!50!white, line width=0.5pt] 
    table [x={Zeit (s)}, y={lowerBound}, col sep=comma]
    {pressure_dat_23-03-07-01.csv};

\addplot[red!50!white, opacity=0.4] 
    fill between[of=upper15 and lower15];

\addplot[name path=upper6, color=blue!50!white, line width=0.5pt] 
    table [x={Zeit (s)}, y={upperBound}, col sep=comma]
    {pressure_dat_23-03-06-05.csv};

\addplot[name path=lower6, color=blue!50!white, line width=0.5pt] 
    table [x={Zeit (s)}, y={lowerBound}, col sep=comma]
    {pressure_dat_23-03-06-05.csv};

\addplot[blue!50!white, opacity=0.4] 
    fill between[of=upper6 and lower6];

\addplot[name path=upper16, color=purple!50!white, line width=0.5pt] 
    table [x={Zeit (s)}, y={upperBound}, col sep=comma]
    {pressure_dat_23-03-07-02.csv};

\addplot[name path=lower16, color=purple!50!white, line width=0.5pt] 
    table [x={Zeit (s)}, y={lowerBound}, col sep=comma]
    {pressure_dat_23-03-07-02.csv};

\addplot[purple!50!white, opacity=0.4] 
    fill between[of=upper16 and lower16];
\end{axis}
\end{tikzpicture}
}

\caption{Comparison of measured (\textbf{M}) and predicted (\textbf{P}) pressure values for the experiments from the Table~\ref{tab:experiments}. (a) shows the increase of operation time with higher cycle number. (b) shows the increase in operation time with different concentrations and similar filter cylce. (c) and (d) show the operational time with different numbers of filter chambers.}
\label{fig:pressure_comparison}
\end{figure}

The model demonstrated a strong reliability for pressure prediction in partially known experiments, as illustrated in Figure~\ref{fig:pressure_comparison} which compares \ac{M} against \ac{P} trends. For experiments with varying filter cycles (Exp-1 to Exp-4, Figure~\ref{fig:pressure_comparison}(a)), the predicted pressure profiles closely matched the measured values, with minor deviations observed at higher pressures during later stages. Similarly, experiments with varying concentrations (Exp-5, Exp-3, and Exp-6, Figure~\ref{fig:pressure_comparison}(b)) showed high accuracy, although Exp-5 exhibited less precision at maximum pressures due to the absence of training data for a concentration of $6.25 g/L$. Figures~\ref{fig:pressure_comparison}(c) and \ref{fig:pressure_comparison}(d) highlight the model's consistency across different configurations, with slight discrepancies near peak pressures.
Quantitative metrics are summarized in Table~\ref{tab:experiments_val}. The errors for partially known experiments ranged from \ac{MSE} $0.006$ to $0.178$, \ac{RMSE} $0.103$ to $0.352$, and RL2N $1.7\%$ to $9.8\%$. The best performance was observed in Exp-5 (RL2N $0.17\%$, PIB $97\%$), while Exp-6 had the highest error (RL2N $9.8\%$, PIB $64\%$). Overall, the model achieved an average \ac{MSE} of $0.048$, \ac{RMSE} of $0.185$, and RL2N of $5\%$, with $82\%$ of the predictions falling within CI90$\%$ bounds and minor deviations for the remaining $18\%$. The pressure drop around $230$ seconds in Exp-6 (Figure~\ref{fig:pressure_comparison}(d)) illustrates the model's ability to follow expected trends despite limited training data for configurations with 3 or 4 plates.

\paragraph{Unknown experiments}

\begin{figure}[h!]

\subfloat[]{
\begin{tikzpicture}
\begin{axis}[
    width=0.5\textwidth, %
    xlabel={Time [s]},
    ylabel={Pressure [bar]},
    legend style={
        at={(0.5,-0.25)}, %
        anchor=north,    
        legend columns=2, %
        /tikz/every even column/.append style={column sep=0.05cm} %
    },
    legend cell align={left}, %
    grid=both, %
    minor grid style={dotted}, %
    major grid style={solid, gray!30} %
]

\addplot[color=blue!70!black, line width=1pt] 
    table [x={Zeit (s)}, y={Druck (bar)}, col sep=comma] 
    {pressure_dat_23-09-11-004.csv};
\addlegendentry{Exp-4 val-M}

\addplot[color=blue!70!black, line width=1pt, dashed] 
    table [x={Zeit (s)}, y={Pred (bar)}, col sep=comma]    
    {pressure_dat_23-09-11-004.csv};
\addlegendentry{Exp-4 val-P}

\addplot[color=green!60!black, line width=1pt] 
    table [x={Zeit (s)}, y={Druck (bar)}, col sep=comma] 
    {pressure_dat_23-09-11-015.csv};
\addlegendentry{Exp-3 val-M}

\addplot[color=green!60!black, line width=1pt, dashed] 
    table [x={Zeit (s)}, y={Pred (bar)}, col sep=comma]    
    {pressure_dat_23-09-11-015.csv};
\addlegendentry{Exp-3 val-P}

\addplot[color=red!70!black, line width=1pt] 
    table [x={Zeit (s)}, y={Druck (bar)}, col sep=comma] 
    {pressure_dat_23-09-11-011.csv};
\addlegendentry{Exp-5 val-M}

\addplot[color=red!70!black, line width=1pt, dashed] 
    table [x={Zeit (s)}, y={Pred (bar)}, col sep=comma]    
    {pressure_dat_23-09-11-011.csv};
\addlegendentry{Exp-5 val-P}

\addplot[name path=upperV, color=red!50!white, line width=0.5pt] 
    table [x={Zeit (s)}, y={upperBound}, col sep=comma]
    {pressure_dat_23-09-11-011.csv};

\addplot[name path=lowerV, color=red!50!white, line width=0.5pt] 
    table [x={Zeit (s)}, y={lowerBound}, col sep=comma]
    {pressure_dat_23-09-11-011.csv};

\addplot[red!50!white, opacity=0.4] 
    fill between[of=upperV and lowerV];

\addplot[name path=upper2V, color=blue!50!white, line width=0.5pt] 
    table [x={Zeit (s)}, y={upperBound}, col sep=comma]
    {pressure_dat_23-09-11-004.csv};

\addplot[name path=lower2V, color=blue!50!white, line width=0.5pt] 
    table [x={Zeit (s)}, y={lowerBound}, col sep=comma]
    {pressure_dat_23-09-11-004.csv};

\addplot[blue!50!white, opacity=0.4] 
    fill between[of=upper2V and lower2V];

\addplot[name path=upper3V, color=green!50!white, line width=0.5pt] 
    table [x={Zeit (s)}, y={upperBound}, col sep=comma]
    {pressure_dat_23-09-11-015.csv};

\addplot[name path=lower3V, color=green!50!white, line width=0.5pt] 
    table [x={Zeit (s)}, y={lowerBound}, col sep=comma]
    {pressure_dat_23-09-11-015.csv};

\addplot[green!50!white, opacity=0.4] 
    fill between[of=upper3V and lower3V];

\end{axis}
\end{tikzpicture}
}
\subfloat[]{
\begin{tikzpicture}
\begin{axis}[
    width=0.5\textwidth, %
    xlabel={Time [s]},
    ylabel={Pressure [bar]},
    legend style={
        at={(0.5,-0.25)}, %
        anchor=north,    
        legend columns=2, %
        /tikz/every even column/.append style={column sep=0.05cm} %
    },
    legend cell align={left}, %
    grid=both, %
    minor grid style={dotted}, %
    major grid style={solid, gray!30} %
]

\addplot[color=blue!70!black, line width=1pt] 
    table [x={Zeit (s)}, y={Druck (bar)}, col sep=comma] 
    {pressure_dat_23-09-11-009.csv};
\addlegendentry{Exp-1 val-M}

\addplot[color=blue!70!black, line width=1pt,dashed] 
    table [x={Zeit (s)}, y={Pred (bar)}, col sep=comma]    
    {pressure_dat_23-09-11-009.csv};
\addlegendentry{Exp-1 val-P}

\addplot[color=green!60!black, line width=1pt] 
    table [x={Zeit (s)}, y={Druck (bar)}, col sep=comma] 
    {pressure_dat_23-09-08-004.csv};
\addlegendentry{Exp-2 val-M}

\addplot[color=green!60!black, line width=1pt,dashed] 
    table [x={Zeit (s)}, y={Pred (bar)}, col sep=comma]    
    {pressure_dat_23-09-08-004.csv};
\addlegendentry{Exp-2 val-P}

 \addplot[color=purple!70!black, line width=1pt, opacity=0] 
     table [x={Druck (bar)}, y={Druck (bar)}, col sep=comma]    
     {pressure_empty.csv}; %
 \addlegendentry{\phantom{Exp}}

\addplot[name path=upper5V, color=blue!50!white, line width=0.5pt] 
    table [x={Zeit (s)}, y={upperBound}, col sep=comma]
    {pressure_dat_23-09-11-009.csv};

\addplot[name path=lower5V, color=blue!50!white, line width=0.5pt] 
    table [x={Zeit (s)}, y={lowerBound}, col sep=comma]
    {pressure_dat_23-09-11-009.csv};

\addplot[blue!50!white, opacity=0.4] 
    fill between[of=upper5V and lower5V];

\addplot[name path=upper6V, color=green!50!white, line width=0.5pt] 
    table [x={Zeit (s)}, y={upperBound}, col sep=comma]
    {pressure_dat_23-09-08-004.csv};

\addplot[name path=lower6V, color=green!50!white, line width=0.5pt] 
    table [x={Zeit (s)}, y={lowerBound}, col sep=comma]
    {pressure_dat_23-09-08-004.csv};

\addplot[green!50!white, opacity=0.4] 
    fill between[of=upper6V and lower6V];

\end{axis}
\end{tikzpicture}
}

\subfloat[]{
\begin{tikzpicture}
\begin{axis}[
    width=0.5\textwidth, %
    xlabel={Time [s]},
    ylabel={Pressure [bar]},
    legend style={
        at={(0.5,-0.25)}, %
        anchor=north,    
        legend columns=2, %
        /tikz/every even column/.append style={column sep=0.05cm} %
    },
    legend cell align={left}, %
    grid=both, %
    minor grid style={dotted}, %
    major grid style={solid, gray!30} %
]

\addplot[color=blue!70!black, line width=1pt] 
    table [x={Zeit (s)}, y={Druck (bar)}, col sep=comma] 
    {pressure_dat_23-09-11-011.csv};
\addlegendentry{Exp-5 val-M}

\addplot[color=blue!70!black, line width=1pt, dashed] 
    table [x={Zeit (s)}, y={Pred (bar)}, col sep=comma]    
    {pressure_dat_23-09-11-011.csv};
\addlegendentry{Exp-5 val-P}

\addplot[color=green!60!black, line width=1pt] 
    table [x={Zeit (s)}, y={Druck (bar)}, col sep=comma] 
    {pressure_dat_23-09-11-003.csv};
\addlegendentry{Exp-6 val-M}

\addplot[color=green!60!black, line width=1pt, dashed] 
    table [x={Zeit (s)}, y={Pred (bar)}, col sep=comma]    
    {pressure_dat_23-09-11-003.csv};
\addlegendentry{Exp-6 val-P}

\addplot[name path=upper8V, color=blue!50!white, line width=0.5pt] 
    table [x={Zeit (s)}, y={upperBound}, col sep=comma]
    {pressure_dat_23-09-11-011.csv};

\addplot[name path=lower8V, color=blue!50!white, line width=0.5pt] 
    table [x={Zeit (s)}, y={lowerBound}, col sep=comma]
    {pressure_dat_23-09-11-011.csv};

\addplot[blue!50!white, opacity=0.4] 
    fill between[of=upper8V and lower8V];

\addplot[name path=upper9V, color=green!50!white, line width=0.5pt] 
    table [x={Zeit (s)}, y={upperBound}, col sep=comma]
    {pressure_dat_23-09-11-003.csv};

\addplot[name path=lower9V, color=green!50!white, line width=0.5pt] 
    table [x={Zeit (s)}, y={lowerBound}, col sep=comma]
    {pressure_dat_23-09-11-003.csv};

\addplot[green!50!white, opacity=0.4] 
    fill between[of=upper9V and lower9V];

\end{axis}
\end{tikzpicture}
}
\hfill
\subfloat[]{
\begin{tikzpicture}
\begin{axis}[
    width=0.5\textwidth, %
    xlabel={Time [s]},
    ylabel={Pressure [bar]},
    legend style={
        at={(0.5,-0.25)}, %
        anchor=north,    
        legend columns=2, %
        /tikz/every even column/.append style={column sep=0.5cm} %
    },
    legend cell align={left}, %
    grid=both, %
    minor grid style={dotted}, %
    major grid style={solid, gray!30} %
]

\addplot[color=blue!70!black, line width=1pt] 
    table [x={Zeit (s)}, y={Druck (bar)}, col sep=comma] 
    {pressure_dat_23-09-08-002.csv};
\addlegendentry{Exp-7 val-M}

\addplot[color=blue!70!black, line width=1pt,dashed] 
    table [x={Zeit (s)}, y={Pred (bar)}, col sep=comma]    
    {pressure_dat_23-09-08-002.csv};
\addlegendentry{Exp-7 val-P}

\addplot[color=green!60!black, line width=1pt] 
    table [x={Zeit (s)}, y={Druck (bar)}, col sep=comma] 
    {pressure_dat_23-09-08-003.csv};
\addlegendentry{Exp-8 val-M}

\addplot[color=green!60!black, line width=1pt,dashed] 
    table [x={Zeit (s)}, y={Pred (bar)}, col sep=comma]    
    {pressure_dat_23-09-08-003.csv};
\addlegendentry{Exp-8 val-P}

\addplot[name path=upper10V, color=blue!50!white, line width=0.5pt] 
    table [x={Zeit (s)}, y={upperBound}, col sep=comma]
    {pressure_dat_23-09-08-002.csv};

\addplot[name path=lower10V, color=blue!50!white, line width=0.5pt] 
    table [x={Zeit (s)}, y={lowerBound}, col sep=comma]
    {pressure_dat_23-09-08-002.csv};

\addplot[blue!50!white, opacity=0.4] 
    fill between[of=upper10V and lower10V];

\addplot[name path=upper11V, color=green!50!white, line width=0.5pt] 
    table [x={Zeit (s)}, y={upperBound}, col sep=comma]
    {pressure_dat_23-09-08-003.csv};

\addplot[name path=lower11V, color=green!50!white, line width=0.5pt] 
    table [x={Zeit (s)}, y={lowerBound}, col sep=comma]
    {pressure_dat_23-09-08-003.csv};

\addplot[green!50!white, opacity=0.4] 
    fill between[of=upper11V and lower11V];

\end{axis}
\end{tikzpicture}
}

\caption{Comparison of measured (\textbf{M}) and predicted (\textbf{P}) pressure values for the experiments listed in Table~\ref{tab:experiments_val}. (a) illustrates the increase in operational time with a higher number of filter cycles. (b) depicts the increase in operational time for different concentrations while maintaining similar filter cycles. (c) demonstrates the variation in operational time with differing numbers of filter chambers. (d) presents the prediction for an unknown concentration of $15 \, \mathrm{g/L}$.}
\label{fig:pressure_comparison_val}
\end{figure}
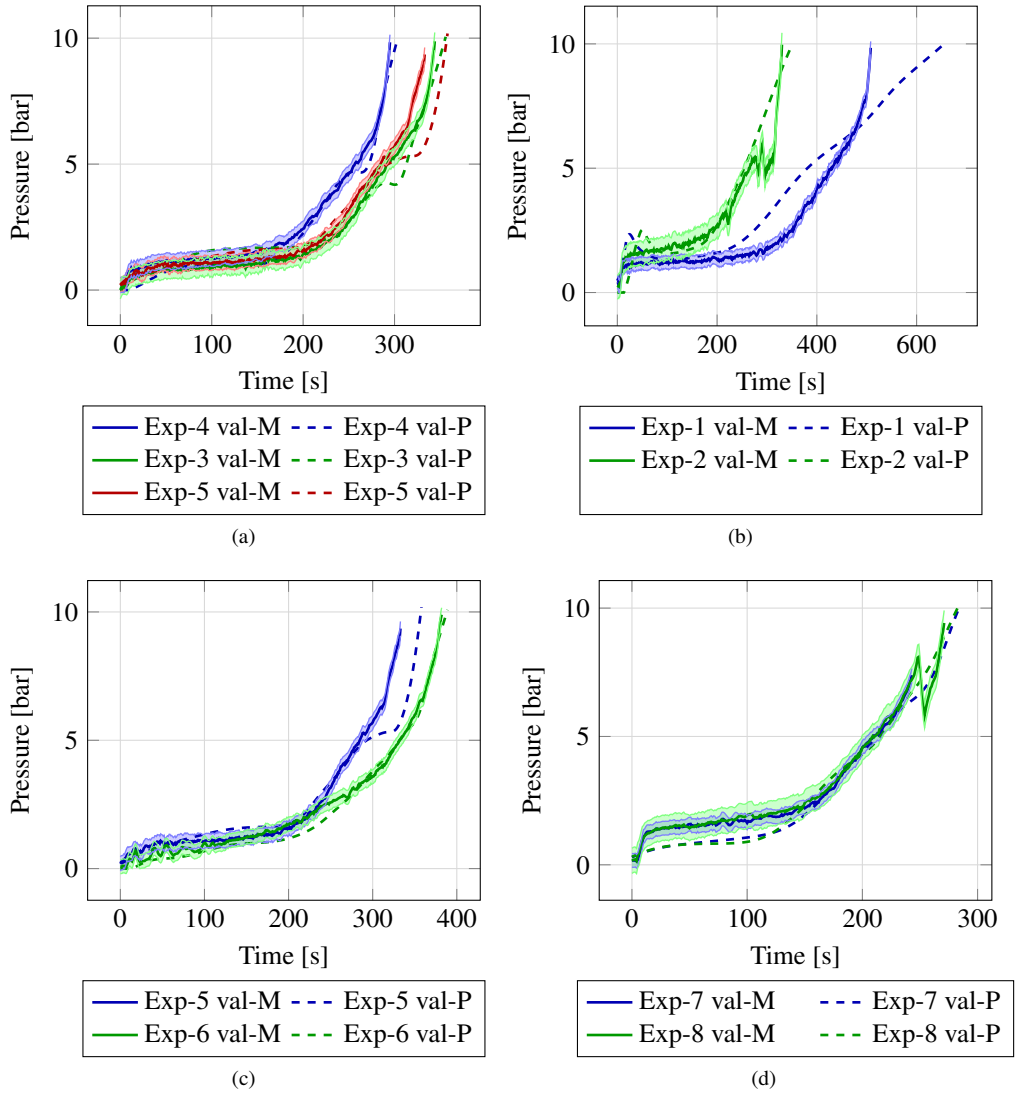

For unknown experiments, the performance of the model was satisfactory, as shown in Figure~\ref{fig:pressure_comparison_val}. The errors for unknown experiments, detailed in Table~\ref{tab:experiments_val}, ranged from \ac{MSE} $0.100$ to $0.838$, \ac{RMSE} $0.317$ to $0.915$, RL2N $9.9\%$ to $28.2\%$, and PIB $64\%$ to $20\%$. Exp-6-val showed the best performance, supported by training data from a similar experiment (Exp-8). In contrast, Exp-1-val exhibited the highest errors due to limited training data at $6.25 g/L$ and narrow filtration cycle ranges (Figure~\ref{fig:pressure_comparison_val}(b)).
Despite these challenges, the model interpolated effectively in experiments with unfamiliar filtration cycles, such as Exp-4-val, achieving good results both qualitatively and quantitatively. Figure~\ref{fig:pressure_comparison_val}(a) highlights deviations at the start and between 260–280 seconds, corresponding to common pressure drops during filtration. Experiments with unknown concentrations (e.g., Exp-7-val and Exp-8-val at $15 g/L$, Figure~\ref{fig:pressure_comparison_val}(c)) further demonstrate the generalization capabilities of the model.
In general, the mean errors for unknown experiments were \ac{MSE} $0.403$, \ac{RMSE} $0.605$, RL2N $18.4\%$, and PIB $49.13\%$, with minimal deviations outside CI90$\%$ bounds (RL2NB $8.2\%$).

\subsubsection{Flow rate prediction}

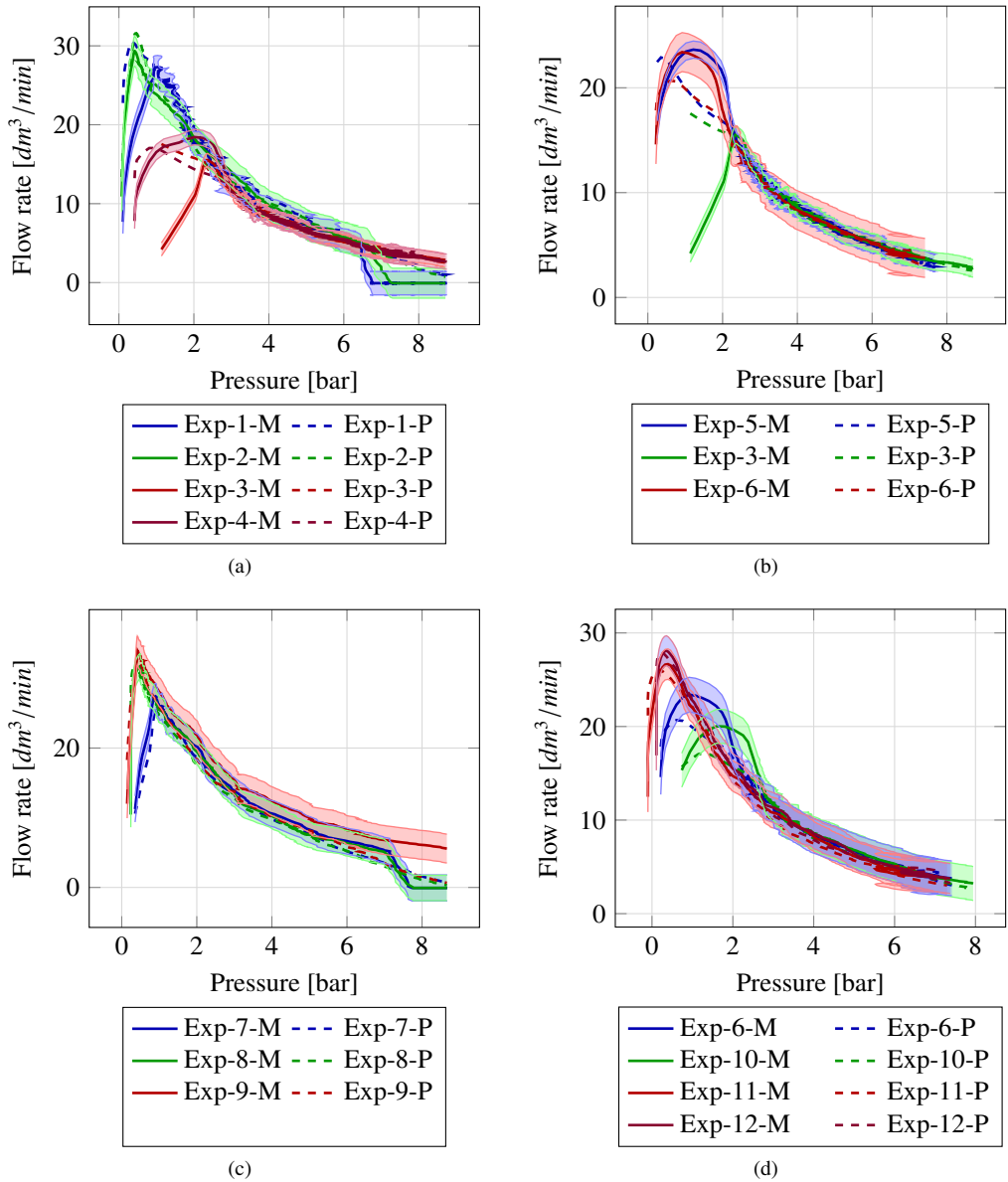
\begin{figure}[h!]
\centering
\subfloat[]{
\begin{tikzpicture}
\begin{axis}[
    width=0.5\textwidth, %
    xlabel={Pressure [bar]},
    ylabel={Flow rate $[dm^3/min]$},
    legend style={
        at={(0.5,-0.25)}, %
        anchor=north,    
        legend columns=2, %
        /tikz/every even column/.append style={column sep=0.05cm} %
    },
    legend cell align={left}, %
    grid=both, %
    minor grid style={dotted}, %
    major grid style={solid, gray!30} %
]

\addplot[color=blue!70!black, line width=1pt] 
    table [x={Druck (bar)}, y={Flow}, col sep=comma] 
    {flow_dat_23-01-16-02.csv};
\addlegendentry{Exp-1-M}

\addplot[color=blue!70!black, line width=1pt,dashed] 
    table [x={Druck (bar)}, y={Flow_pred}, col sep=comma]    
    {flow_dat_23-01-16-02.csv};
\addlegendentry{Exp-1-P}

\addplot[color=green!60!black, line width=1pt] 
    table [x={Druck (bar)}, y={Flow}, col sep=comma] 
    {flow_dat_23-02-20-01.csv};
\addlegendentry{Exp-2-M}

\addplot[color=green!60!black, line width=1pt,dashed] 
    table [x={Druck (bar)}, y={Flow_pred}, col sep=comma]    
    {flow_dat_23-02-20-01.csv};
\addlegendentry{Exp-2-P}

\addplot[color=red!70!black, line width=1pt] 
    table [x={Druck (bar)}, y={Flow}, col sep=comma] 
    {flow_dat_23-03-14-01.csv};
\addlegendentry{Exp-3-M}

\addplot[color=red!70!black, line width=1pt,dashed] 
    table [x={Druck (bar)}, y={Flow_pred}, col sep=comma]    
    {flow_dat_23-03-14-01.csv};
\addlegendentry{Exp-3-P}

\addplot[color=purple!70!black, line width=1pt] 
    table [x={Druck (bar)}, y={Flow}, col sep=comma] 
    {flow_dat_23-03-14-02.csv};
\addlegendentry{Exp-4-M}

\addplot[color=purple!70!black, line width=1pt,dashed] 
    table [x={Druck (bar)}, y={Flow_pred}, col sep=comma]    
    {flow_dat_23-03-14-02.csv};
\addlegendentry{Exp-4-P}

\addplot[name path=upperF, color=blue!50!white, line width=0.5pt] 
    table [x={Druck (bar)}, y={upperBound}, col sep=comma]
    {flow_dat_23-01-16-02.csv};

\addplot[name path=lowerF, color=blue!50!white, line width=0.5pt] 
    table [x={Druck (bar)}, y={lowerBound}, col sep=comma]
    {flow_dat_23-01-16-02.csv};

\addplot[blue!50!white, opacity=0.4] 
    fill between[of=upperF and lowerF];

\addplot[name path=upper2F, color=green!50!white, line width=0.5pt] 
    table [x={Druck (bar)}, y={upperBound}, col sep=comma]
    {flow_dat_23-02-20-01.csv};

\addplot[name path=lower2F, color=green!50!white, line width=0.5pt] 
    table [x={Druck (bar)}, y={lowerBound}, col sep=comma]
    {flow_dat_23-02-20-01.csv};

\addplot[green!50!white, opacity=0.4] 
    fill between[of=upper2F and lower2F];

\addplot[name path=upper3F, color=red!50!white, line width=0.5pt] 
    table [x={Druck (bar)}, y={upperBound}, col sep=comma]
    {flow_dat_23-03-14-01.csv};

\addplot[name path=lower3F, color=red!50!white, line width=0.5pt] 
    table [x={Druck (bar)}, y={lowerBound}, col sep=comma]
    {flow_dat_23-03-14-01.csv};

\addplot[red!50!white, opacity=0.4] 
    fill between[of=upper3F and lower3F];

\addplot[name path=upper4F, color=purple!50!white, line width=0.5pt] 
    table [x={Druck (bar)}, y={upperBound}, col sep=comma]
    {flow_dat_23-03-14-02.csv};

\addplot[name path=lower4F, color=purple!50!white, line width=0.5pt] 
    table [x={Druck (bar)}, y={lowerBound}, col sep=comma]
    {flow_dat_23-03-14-02.csv};

\addplot[purple!50!white, opacity=0.4] 
    fill between[of=upper4F and lower4F];

\end{axis}
\end{tikzpicture}
}
\hfill
\subfloat[]{

\begin{tikzpicture}
\begin{axis}[
    width=0.5\textwidth, %
    xlabel={Pressure [bar]},
    ylabel={Flow rate $[dm^3/min]$},
    legend style={
        at={(0.5,-0.25)}, %
        anchor=north,    
        legend columns=2, %
        /tikz/every even column/.append style={column sep=0.5cm} %
    },
    legend cell align={left}, %
    grid=both, %
    minor grid style={dotted}, %
    major grid style={solid, gray!30} %
]

\addplot[color=blue!70!black, line width=1pt] 
    table [x={Druck (bar)}, y={Flow}, col sep=comma] 
    {flow_dat_23-03-07-05.csv};
\addlegendentry{Exp-5-M}

\addplot[color=blue!70!black, line width=1pt,dashed] 
    table [x={Druck (bar)}, y={Flow_pred}, col sep=comma]    
    {flow_dat_23-03-07-05.csv};
\addlegendentry{Exp-5-P}

\addplot[color=green!60!black, line width=1pt] 
    table [x={Druck (bar)}, y={Flow}, col sep=comma] 
    {flow_dat_23-03-14-01.csv};
\addlegendentry{Exp-3-M}

\addplot[color=green!60!black, line width=1pt,dashed] 
    table [x={Druck (bar)}, y={Flow_pred}, col sep=comma]    
    {flow_dat_23-03-14-01.csv};
\addlegendentry{Exp-3-P}

\addplot[color=red!70!black, line width=1pt] 
    table [x={Druck (bar)}, y={Flow}, col sep=comma] 
    {flow_dat_23-03-06-05.csv};
\addlegendentry{Exp-6-M}

\addplot[color=red!70!black, line width=1pt,dashed] 
    table [x={Druck (bar)}, y={Flow_pred}, col sep=comma]    
    {flow_dat_23-03-06-05.csv};
\addlegendentry{Exp-6-P}

\addplot[color=purple!70!black, line width=1pt, opacity=0] 
    table [x={Druck (bar)}, y={Druck (bar)}, col sep=comma]    
    {pressure_empty.csv}; %
\addlegendentry{\phantom{Exp}}

\addplot[name path=upper5F, color=blue!50!white, line width=0.5pt] 
    table [x={Druck (bar)}, y={upperBound}, col sep=comma]
    {flow_dat_23-03-07-05.csv};

\addplot[name path=lower5F, color=blue!50!white, line width=0.5pt] 
    table [x={Druck (bar)}, y={lowerBound}, col sep=comma]
    {flow_dat_23-03-07-05.csv};

\addplot[blue!50!white, opacity=0.4] 
    fill between[of=upper5F and lower5F];

\addplot[name path=upper6F, color=green!50!white, line width=0.5pt] 
    table [x={Druck (bar)}, y={upperBound}, col sep=comma]
    {flow_dat_23-03-14-01.csv};

\addplot[name path=lower6F, color=green!50!white, line width=0.5pt] 
    table [x={Druck (bar)}, y={lowerBound}, col sep=comma]
    {flow_dat_23-03-14-01.csv};

\addplot[green!50!white, opacity=0.4] 
    fill between[of=upper6F and lower6F];

\addplot[name path=upper7F, color=red!50!white, line width=0.5pt] 
    table [x={Druck (bar)}, y={upperBound}, col sep=comma]
    {flow_dat_23-03-06-05.csv};

\addplot[name path=lower7F, color=red!50!white, line width=0.5pt] 
    table [x={Druck (bar)}, y={lowerBound}, col sep=comma]
    {flow_dat_23-03-06-05.csv};

\addplot[red!50!white, opacity=0.4] 
    fill between[of=upper7F and lower7F];

\end{axis}
\end{tikzpicture}
}
\\
\subfloat[]{
\begin{tikzpicture}
\begin{axis}[
    width=0.5\textwidth, %
    xlabel={Pressure [bar]},
    ylabel={Flow rate $[dm^3/min]$},
    legend style={
        at={(0.5,-0.25)}, %
        anchor=north,    
        legend columns=2, %
        /tikz/every even column/.append style={column sep=0.05cm} %
    },
    legend cell align={left}, %
    grid=both, %
    minor grid style={dotted}, %
    major grid style={solid, gray!30} %
]

\addplot[color=blue!70!black, line width=1pt] 
    table [x={Druck (bar)}, y={Flow}, col sep=comma] 
    {flow_dat_23-02-20-05.csv};
\addlegendentry{Exp-7-M}

\addplot[color=blue!70!black, line width=1pt, dashed] 
    table [x={Druck (bar)}, y={Flow_pred}, col sep=comma]    
    {flow_dat_23-02-20-05.csv};
\addlegendentry{Exp-7-P}

\addplot[color=green!60!black, line width=1pt] 
    table [x={Druck (bar)}, y={Flow}, col sep=comma] 
    {flow_dat_23-02-20-04.csv};
\addlegendentry{Exp-8-M}

\addplot[color=green!60!black, line width=1pt,dashed] 
    table [x={Druck (bar)}, y={Flow_pred}, col sep=comma]    
    {flow_dat_23-02-20-04.csv};
\addlegendentry{Exp-8-P}

\addplot[color=red!70!black, line width=1pt] 
    table [x={Druck (bar)}, y={Flow}, col sep=comma] 
    {flow_dat_23-02-17-03.csv};
\addlegendentry{Exp-9-M}

\addplot[color=red!70!black, line width=1pt,dashed] 
    table [x={Druck (bar)}, y={Flow_pred}, col sep=comma]    
    {flow_dat_23-02-17-03.csv};
\addlegendentry{Exp-9-P}

\addplot[color=purple!70!black, line width=1pt, opacity=0] 
    table [x={Druck (bar)}, y={Druck (bar)}, col sep=comma]    
    {pressure_empty.csv}; %
\addlegendentry{\phantom{Exp}}

\addplot[name path=upper8F, color=blue!50!white, line width=0.5pt] 
    table [x={Druck (bar)}, y={upperBound}, col sep=comma]
    {flow_dat_23-02-20-05.csv};

\addplot[name path=lower8F, color=blue!50!white, line width=0.5pt] 
    table [x={Druck (bar)}, y={lowerBound}, col sep=comma]
    {flow_dat_23-02-20-05.csv};

\addplot[blue!50!white, opacity=0.4] 
    fill between[of=upper8F and lower8F];

\addplot[name path=upper9F, color=green!50!white, line width=0.5pt] 
    table [x={Druck (bar)}, y={upperBound}, col sep=comma]
    {flow_dat_23-02-20-04.csv};

\addplot[name path=lower9F, color=green!50!white, line width=0.5pt] 
    table [x={Druck (bar)}, y={lowerBound}, col sep=comma]
    {flow_dat_23-02-20-04.csv};

\addplot[green!50!white, opacity=0.4] 
    fill between[of=upper9F and lower9F];

\addplot[name path=upper10F, color=red!50!white, line width=0.5pt] 
    table [x={Druck (bar)}, y={upperBound}, col sep=comma]
    {flow_dat_23-02-17-03.csv};

\addplot[name path=lower10F, color=red!50!white, line width=0.5pt] 
    table [x={Druck (bar)}, y={lowerBound}, col sep=comma]
    {flow_dat_23-02-17-03.csv};

\addplot[red!50!white, opacity=0.4] 
    fill between[of=upper10F and lower10F];

\end{axis}
\end{tikzpicture}
}
\hfill
\subfloat[]{
\begin{tikzpicture}
\begin{axis}[
    width=0.5\textwidth, %
    xlabel={Pressure [bar]},
    ylabel={Flow rate $[dm^3/min]$},
    legend style={
        at={(0.5,-0.25)}, %
        anchor=north,    
        legend columns=2, %
        /tikz/every even column/.append style={column sep=0.5cm} %
    },
    legend cell align={left}, %
    grid=both, %
    minor grid style={dotted}, %
    major grid style={solid, gray!30} %
]

\addplot[color=blue!70!black, line width=1pt] 
    table [x={Druck (bar)}, y={Flow}, col sep=comma] 
    {flow_dat_23-03-06-05.csv};
\addlegendentry{Exp-6-M}

\addplot[color=blue!70!black, line width=1pt,dashed] 
    table [x={Druck (bar)}, y={Flow_pred}, col sep=comma]    
    {flow_dat_23-03-06-05.csv};
\addlegendentry{Exp-6-P}

\addplot[color=green!60!black, line width=1pt] 
    table [x={Druck (bar)}, y={Flow}, col sep=comma] 
    {flow_dat_23-03-06-06.csv};
\addlegendentry{Exp-10-M}

\addplot[color=green!60!black, line width=1pt,dashed] 
    table [x={Druck (bar)}, y={Flow_pred}, col sep=comma]    
    {flow_dat_23-03-06-06.csv};
\addlegendentry{Exp-10-P}

\addplot[color=red!70!black,  line width=1pt] 
    table [x={Druck (bar)}, y={Flow}, col sep=comma] 
    {flow_dat_23-03-07-01.csv};
\addlegendentry{Exp-11-M}

\addplot[color=red!70!black, line width=1pt,dashed] 
    table [x={Druck (bar)}, y={Flow_pred}, col sep=comma]    
    {flow_dat_23-03-07-01.csv};
\addlegendentry{Exp-11-P}

\addplot[color=purple!70!black, line width=1pt] 
    table [x={Druck (bar)}, y={Flow}, col sep=comma] 
    {flow_dat_23-03-07-02.csv};
\addlegendentry{Exp-12-M}

\addplot[color=purple!70!black, line width=1pt,dashed] 
    table [x={Druck (bar)}, y={Flow_pred}, col sep=comma]    
    {flow_dat_23-03-07-02.csv};
\addlegendentry{Exp-12-P}

\addplot[name path=upper11F, color=green!50!white, line width=0.5pt] 
    table [x={Druck (bar)}, y={upperBound}, col sep=comma]
    {flow_dat_23-03-06-06.csv};

\addplot[name path=lower11F, color=green!50!white, line width=0.5pt] 
    table [x={Druck (bar)}, y={lowerBound}, col sep=comma]
    {flow_dat_23-03-06-06.csv};

\addplot[green!50!white, opacity=0.4] 
    fill between[of=upper11F and lower11F];

\addplot[name path=upper12F, color=red!50!white, line width=0.5pt] 
    table [x={Druck (bar)}, y={upperBound}, col sep=comma]
    {flow_dat_23-03-07-01.csv};

\addplot[name path=lower12F, color=red!50!white, line width=0.5pt] 
    table [x={Druck (bar)}, y={lowerBound}, col sep=comma]
    {flow_dat_23-03-07-01.csv};

\addplot[red!50!white, opacity=0.4] 
    fill between[of=upper12F and lower12F];

\addplot[name path=upper13F, color=blue!50!white, line width=0.5pt] 
    table [x={Druck (bar)}, y={upperBound}, col sep=comma]
    {flow_dat_23-03-06-05.csv};

\addplot[name path=lower13F, color=blue!50!white, line width=0.5pt] 
    table [x={Druck (bar)}, y={lowerBound}, col sep=comma]
    {flow_dat_23-03-06-05.csv};

\addplot[blue!50!white, opacity=0.4] 
    fill between[of=upper13F and lower13F];

\addplot[name path=upper14F, color=purple!50!white, line width=0.5pt] 
    table [x={Druck (bar)}, y={upperBound}, col sep=comma]
    {flow_dat_23-03-07-02.csv};

\addplot[name path=lower14F, color=purple!50!white, line width=0.5pt] 
    table [x={Druck (bar)}, y={lowerBound}, col sep=comma]
    {flow_dat_23-03-07-02.csv};
\addplot[blue!50!white, opacity=0.4] 
    fill between[of=upper14F and lower14F];

\end{axis}
\end{tikzpicture}
}
\caption{Comparison of measured (\textbf{M}) and predicted (\textbf{P}) flow rate for the experiments from the Table~\ref{tab:experiments}. (a) shows the reduction of the flow rate with higher cycle number over the pressure. (b) shows the reduction of the flow rate with different concentrations and similar filter cylces over the pressure. (c) and (d) shows the flow rate with different numbers of filter chambers over the pressure.}
\label{fig:flow_comparison}
\end{figure}

The model performed well for flow rate predictions, closely aligning with the experimental setups, as illustrated in Figure~\ref{fig:flow_comparison}. The maximum flow rates showed expected trends, such as declines with increasing filter cycles (Figure~\ref{fig:flow_comparison}(a)) and concentrations (Figure~\ref{fig:flow_comparison}(b)). The predictions for varying filter chamber numbers (Figures~\ref{fig:flow_comparison}(c) and \ref{fig:flow_comparison}(d)) also matched measured values qualitatively.
Quantitative errors for partially known experiments, summarized in Table~\ref{tab:experiments_val}, ranged from \ac{MSE} $0.339$ to $2.150$, \ac{RMSE} $0.582$ to $1.466$, and RL2N $6.2\%$ to $16.7\%$, with $77\%$ to $89\%$ of predictions within CI90$\%$ bounds. Deviations were linked to specific anomalies. For example, Exp-3 showed initial discrepancies due to clogging, which normalized post-clogging. Exp-5 achieved the best results, closely followed by Exp-12. As seen in Figure~\ref{fig:flow_comparison}(b), the significant deviation of Exp-5 at the start reflects challenges in initial phase predictions.

\paragraph{Unknown experiments}

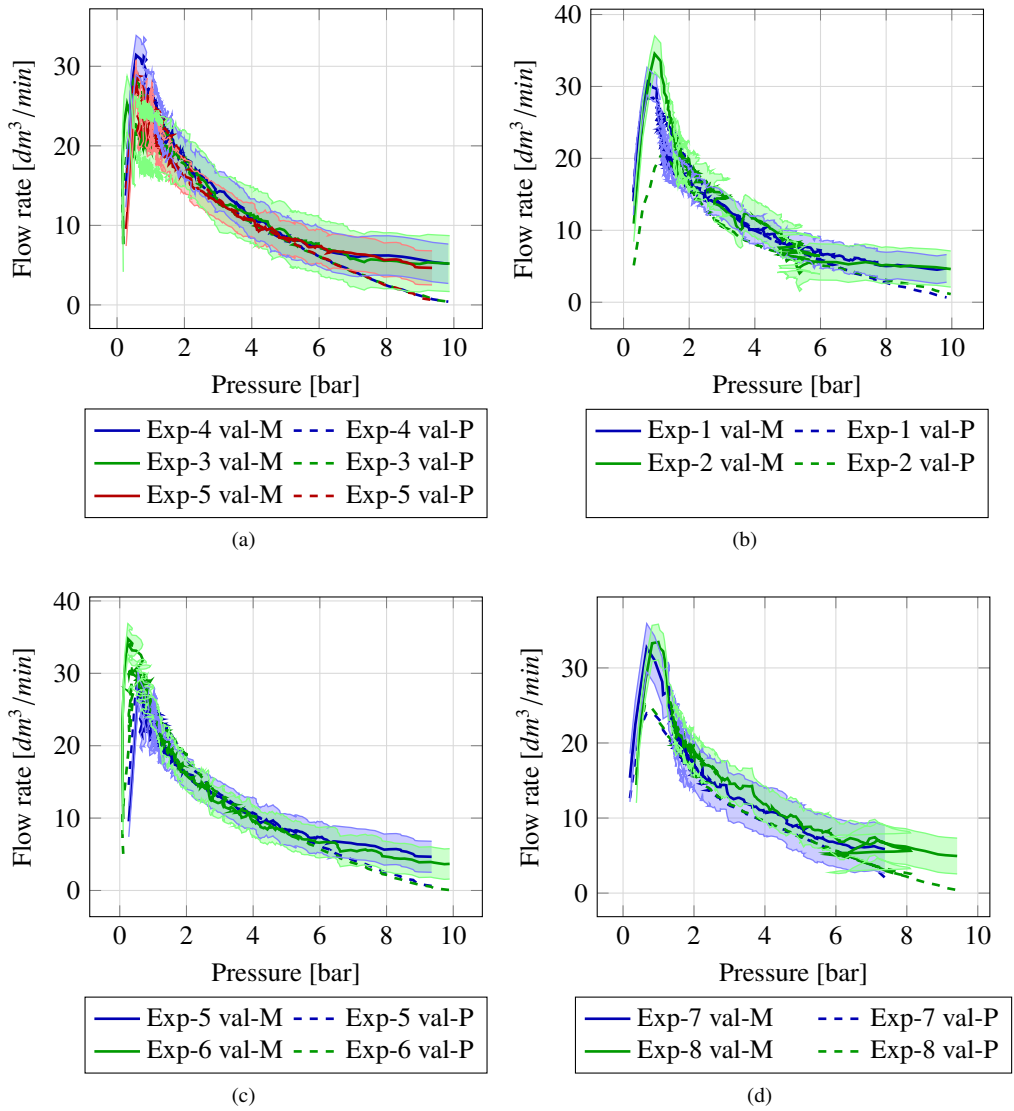
\begin{figure}[h!]

\subfloat[]{
\begin{tikzpicture}
\begin{axis}[
    width=0.5\textwidth, %
    xlabel={Pressure [bar]},
    ylabel={Flow rate [$dm^3/min]$},
    legend style={
        at={(0.5,-0.25)}, %
        anchor=north,    
        legend columns=2, %
        /tikz/every even column/.append style={column sep=0.05cm} %
    },
    legend cell align={left}, %
    grid=both, %
    minor grid style={dotted}, %
    major grid style={solid, gray!30} %
]

\addplot[color=blue!70!black, line width=1pt] 
    table [x={Druck (bar)}, y={Flow}, col sep=comma] 
    {flow_dat_23-09-11-004.csv};
\addlegendentry{Exp-4 val-M}

\addplot[color=blue!70!black, line width=1pt, dashed] 
    table [x={Druck (bar)}, y={Flow_pred}, col sep=comma]    
    {flow_dat_23-09-11-004.csv};
\addlegendentry{Exp-4 val-P}

\addplot[color=green!60!black, line width=1pt] 
    table [x={Druck (bar)}, y={Flow}, col sep=comma] 
    {flow_dat_23-09-11-015.csv};
\addlegendentry{Exp-3 val-M}

\addplot[color=green!60!black, line width=1pt, dashed] 
    table [x={Druck (bar)}, y={Flow_pred}, col sep=comma]    
    {flow_dat_23-09-11-015.csv};
\addlegendentry{Exp-3 val-P}

\addplot[color=red!70!black, line width=1pt] 
    table [x={Druck (bar)}, y={Flow}, col sep=comma] 
    {flow_dat_23-09-11-011.csv};
\addlegendentry{Exp-5 val-M}

\addplot[color=red!70!black, line width=1pt, dashed] 
    table [x={Druck (bar)}, y={Flow_pred}, col sep=comma]    
    {flow_dat_23-09-11-011.csv};
\addlegendentry{Exp-5 val-P}

\addplot[name path=upperFV, color=red!50!white, line width=0.5pt] 
    table [x={Druck (bar)}, y={upperBound}, col sep=comma]
    {flow_dat_23-09-11-011.csv};

\addplot[name path=lowerFV, color=red!50!white, line width=0.5pt] 
    table [x={Druck (bar)}, y={lowerBound}, col sep=comma]
    {flow_dat_23-09-11-011.csv};

\addplot[red!50!white, opacity=0.4] 
    fill between[of=upperFV and lowerFV];

\addplot[name path=upperF2V, color=blue!50!white, line width=0.5pt] 
    table [x={Druck (bar)}, y={upperBound}, col sep=comma]
    {flow_dat_23-09-11-004.csv};

\addplot[name path=lowerF2V, color=blue!50!white, line width=0.5pt] 
    table [x={Druck (bar)}, y={lowerBound}, col sep=comma]
    {flow_dat_23-09-11-004.csv};

\addplot[blue!50!white, opacity=0.4] 
    fill between[of=upperF2V and lowerF2V];

\addplot[name path=upperF3V, color=green!50!white, line width=0.5pt] 
    table [x={Druck (bar)}, y={upperBound}, col sep=comma]
    {flow_dat_23-09-11-015.csv};

\addplot[name path=lowerF3V, color=green!50!white, line width=0.5pt] 
    table [x={Druck (bar)}, y={lowerBound}, col sep=comma]
    {flow_dat_23-09-11-015.csv};

\addplot[green!50!white, opacity=0.4] 
    fill between[of=upperF3V and lowerF3V];

\end{axis}
\end{tikzpicture}
}
\subfloat[]{
\begin{tikzpicture}
\begin{axis}[
    width=0.5\textwidth, %
    xlabel={Pressure [bar]},
    ylabel={Flow rate [$dm^3/min]$},
    legend style={
        at={(0.5,-0.25)}, %
        anchor=north,    
        legend columns=2, %
        /tikz/every even column/.append style={column sep=0.05cm} %
    },
    legend cell align={left}, %
    grid=both, %
    minor grid style={dotted}, %
    major grid style={solid, gray!30} %
]

\addplot[color=blue!70!black, line width=1pt] 
    table [x={Druck (bar)}, y={Flow}, col sep=comma] 
    {flow_dat_23-09-11-009.csv};
\addlegendentry{Exp-1 val-M}

\addplot[color=blue!70!black, line width=1pt,dashed] 
    table [x={Druck (bar)}, y={Flow_pred}, col sep=comma]    
    {flow_dat_23-09-11-009.csv};
\addlegendentry{Exp-1 val-P}

\addplot[color=green!60!black, line width=1pt] 
    table [x={Druck (bar)}, y={Flow}, col sep=comma] 
    {flow_dat_23-09-08-004.csv};
\addlegendentry{Exp-2 val-M}

\addplot[color=green!60!black, line width=1pt,dashed] 
    table [x={Druck (bar)}, y={Flow_pred}, col sep=comma]    
    {flow_dat_23-09-08-004.csv};
\addlegendentry{Exp-2 val-P}

 \addplot[color=purple!70!black, line width=1pt, opacity=0] 
     table [x={Druck (bar)}, y={Druck (bar)}, col sep=comma]    
     {pressure_empty.csv}; %
 \addlegendentry{\phantom{Exp}}

\addplot[name path=upperF5V, color=blue!50!white, line width=0.5pt] 
    table [x={Druck (bar)}, y={upperBound}, col sep=comma]
    {flow_dat_23-09-11-009.csv};

\addplot[name path=lowerF5V, color=blue!50!white, line width=0.5pt] 
    table [x={Druck (bar)}, y={lowerBound}, col sep=comma]
    {flow_dat_23-09-11-009.csv};

\addplot[blue!50!white, opacity=0.4] 
    fill between[of=upperF5V and lowerF5V];

\addplot[name path=upperF6V, color=green!50!white, line width=0.5pt] 
    table [x={Druck (bar)}, y={upperBound}, col sep=comma]
    {flow_dat_23-09-08-004.csv};

\addplot[name path=lowerF6V, color=green!50!white, line width=0.5pt] 
    table [x={Druck (bar)}, y={lowerBound}, col sep=comma]
    {flow_dat_23-09-08-004.csv};

\addplot[green!50!white, opacity=0.4] 
    fill between[of=upperF6V and lowerF6V];

\end{axis}
\end{tikzpicture}
}

\subfloat[]{
\begin{tikzpicture}
\begin{axis}[
    width=0.5\textwidth, %
    xlabel={Pressure [bar]},
    ylabel={Flow rate [$dm^3/min]$},
    legend style={
        at={(0.5,-0.25)}, %
        anchor=north,    
        legend columns=2, %
        /tikz/every even column/.append style={column sep=0.05cm} %
    },
    legend cell align={left}, %
    grid=both, %
    minor grid style={dotted}, %
    major grid style={solid, gray!30} %
]

\addplot[color=blue!70!black, line width=1pt] 
    table [x={Druck (bar)}, y={Flow}, col sep=comma] 
    {flow_dat_23-09-11-011.csv};
\addlegendentry{Exp-5 val-M}

\addplot[color=blue!70!black, line width=1pt, dashed] 
    table [x={Druck (bar)}, y={Flow_pred}, col sep=comma]    
    {flow_dat_23-09-11-011.csv};
\addlegendentry{Exp-5 val-P}

\addplot[color=green!60!black, line width=1pt] 
    table [x={Druck (bar)}, y={Flow}, col sep=comma] 
    {flow_dat_23-09-11-003.csv};
\addlegendentry{Exp-6 val-M}

\addplot[color=green!60!black, line width=1pt, dashed] 
    table [x={Druck (bar)}, y={Flow_pred}, col sep=comma]    
    {flow_dat_23-09-11-003.csv};
\addlegendentry{Exp-6 val-P}

\addplot[name path=upperF8V, color=blue!50!white, line width=0.5pt] 
    table [x={Druck (bar)}, y={upperBound}, col sep=comma]
    {flow_dat_23-09-11-011.csv};

\addplot[name path=lowerF8V, color=blue!50!white, line width=0.5pt] 
    table [x={Druck (bar)}, y={lowerBound}, col sep=comma]
    {flow_dat_23-09-11-011.csv};

\addplot[blue!50!white, opacity=0.4] 
    fill between[of=upperF8V and lowerF8V];

\addplot[name path=upperF9V, color=green!50!white, line width=0.5pt] 
    table [x={Druck (bar)}, y={upperBound}, col sep=comma]
    {flow_dat_23-09-11-003.csv};

\addplot[name path=lowerF9V, color=green!50!white, line width=0.5pt] 
    table [x={Druck (bar)}, y={lowerBound}, col sep=comma]
    {flow_dat_23-09-11-003.csv};

\addplot[green!50!white, opacity=0.4] 
    fill between[of=upperF9V and lowerF9V];

\end{axis}
\end{tikzpicture}
}
\hfill
\subfloat[]{
\begin{tikzpicture}
\begin{axis}[
    width=0.5\textwidth, %
    xlabel={Pressure [bar]},
    ylabel={Flow rate [$dm^3/min]$},
    legend style={
        at={(0.5,-0.25)}, %
        anchor=north,    
        legend columns=2, %
        /tikz/every even column/.append style={column sep=0.5cm} %
    },
    legend cell align={left}, %
    grid=both, %
    minor grid style={dotted}, %
    major grid style={solid, gray!30} %
]

\addplot[color=blue!70!black, line width=1pt] 
    table [x={Druck (bar)}, y={Flow}, col sep=comma] 
    {flow_dat_23-09-08-002.csv};
\addlegendentry{Exp-7 val-M}

\addplot[color=blue!70!black, line width=1pt, dashed] 
    table [x={Druck (bar)}, y={Flow_pred}, col sep=comma]    
    {flow_dat_23-09-08-002.csv};
\addlegendentry{Exp-7 val-P}

\addplot[color=green!60!black, line width=1pt] 
    table [x={Druck (bar)}, y={Flow}, col sep=comma] 
    {flow_dat_23-09-08-003.csv};
\addlegendentry{Exp-8 val-M}

\addplot[color=green!60!black, line width=1pt,dashed] 
    table [x={Druck (bar)}, y={Flow_pred}, col sep=comma]    
    {flow_dat_23-09-08-003.csv};
\addlegendentry{Exp-8 val-P}

\addplot[name path=upperF10V, color=blue!50!white, line width=0.5pt] 
    table [x={Druck (bar)}, y={upperBound}, col sep=comma]
    {flow_dat_23-09-08-002.csv};

\addplot[name path=lowerF10V, color=blue!50!white, line width=0.5pt] 
    table [x={Druck (bar)}, y={lowerBound}, col sep=comma]
    {flow_dat_23-09-08-002.csv};

\addplot[blue!50!white, opacity=0.4] 
    fill between[of=upperF10V and lowerF10V];

\addplot[name path=upperF11V, color=green!50!white, line width=0.5pt] 
    table [x={Druck (bar)}, y={upperBound}, col sep=comma]
    {flow_dat_23-09-08-003.csv};

\addplot[name path=lowerF11V, color=green!50!white, line width=0.5pt] 
    table [x={Druck (bar)}, y={lowerBound}, col sep=comma]
    {flow_dat_23-09-08-003.csv};

\addplot[green!50!white, opacity=0.4] 
    fill between[of=upperF11V and lowerF11V];

\end{axis}
\end{tikzpicture}
}

\caption{Comparison of measured (\textbf{M}) and predicted (\textbf{P}) flow rate for the experiments listed in Table~\ref{tab:experiments_val}. (a) illustrates the reduction of flow rate with a higher number of filter cycles. (b)  shows the reduction of the flow rate with
different concentrations and similar filter cylces over the pressure. (c) shows the increase in flow rate with different number of filter chambers. (d) presents the prediction of the flow rate for an unknown concentration of $15 \, \mathrm{g/L}$.}
\label{fig:flow_comparison_val}
\end{figure}

For unknown experiments, the prediction errors, detailed in Table~\ref{tab:error_only_val}, were higher, with averages of \ac{MSE} $8.229$, \ac{RMSE} $2.772$, RL2N $15.4\%$, and PIB $52.25\%$. The worst predictions were in Exp-2-val, where initial trends deviated strongly (Figure~\ref{fig:flow_comparison_val}(a) and \ref{fig:flow_comparison_val}(b)) due to limited training data for higher flow rates ($7 \text{dm}^3/\text{min}$). In contrast, Exp-6-val performed best, with deviations confined to the end of the filtration process, caused by sensor limitations below $5~\text{dm}^3/\text{min}$.
Experiments at unknown concentrations ($15 g/L$, Exp-2-val and Exp-3-val) achieved acceptable results, with \ac{MSE} $4.876$ to $9.649$, \ac{RMSE} $2.208$ to $3.106$, RL2N $12.7\%$ to $17.9\%$, and PIB $63\%$ to $45\%$. Figure~\ref{fig:flow_comparison_val}(c) illustrates the model's ability to approximate overall trends despite missing knowledge for certain configurations.

\begin{landscape}  %

\begin{table}[h!]
\centering
\caption{Prediction errors for experiments in the validation and training sets, evaluated for both pressure and flow rate. Metrics include: MSE (mean squared error), RMSE (root mean squared error), RL2N (relative $L^2$ norm error), RL2N-B (relative $L^2$ norm error with respect to CI90 bounds), and PIB (percentage of points within bounds).}
\label{tabel:error_train_val}
\resizebox{\textwidth}{!}{  %
\begin{tabular}{c|ccccc|ccccc}
\toprule
\multirow{2.5}{*}{\textbf{Experiment}} & \multicolumn{5}{c|}{\textbf{Pressure}} & \multicolumn{5}{c}{\textbf{Flow rate}} \\
                    & \textbf{MSE} & \textbf{RMSE} & \textbf{RL2N} [$\%$]& \textbf{RL2N-B} [$\%$]& \textbf{PIB [\%]} & \textbf{MSE} & \textbf{RMSE} & \textbf{RL2N} [$\%$]& \textbf{RL2N-B} [$\%$]& \textbf{PIB [$\%$]} \\
\midrule
1 & 0.041 & 0.202 & 4.6 & 0.5 & 96 & 4.955 & 2.226 & 13.4 & 5.3 & 50 \\ 
2 & 0.011 & 0.103 & 2.6 & 0.1 & 98 & 3.868 & 1.967 & 9.8 & 2.3 & 64 \\ 
3 & 0.041 & 0.202 & 4.1 & 1.3 & 74 & 2.150 & 1.466 & 16.7 & 10.6 & 77 \\ 
4 & 0.036 & 0.190 & 3.7 & 0.9 & 75 & 0.468 & 0.684 & 8.4 & 3.4 & 69 \\ 
5 & 0.006 & 0.075 & 1.7 & 0.3 & 97 & 0.339 & 0.582 & 6.2 & 2.6 & 89 \\ 
6 & 0.178 & 0.421 & 9.8 & 4.7 & 64 & 1.413 & 1.189 & 10.2 & 3.6 & 62 \\ 
7 & 0.013 & 0.114 & 2.7 & 0.2 & 98 & 2.905 & 1.704 & 9.2 & 3.3 & 60 \\ 
8 & 0.045 & 0.211 & 5.8 & 2.4 & 76 & 3.953 & 1.988 & 9.1 & 4.9 & 91 \\ 
9 & 0.008 & 0.088 & 2.2 & 0.0 & 100 & 3.934 & 1.983 & 8.5 & 1.6 & 77 \\ 
10 & 0.038 & 0.196 & 4.1 & 0.5 & 86 & 0.984 & 0.992 & 9.9 & 2.7 & 66 \\ 
11 & 0.066 & 0.257 & 6.7 & 2.3 & 75 & 1.082 & 1.040 & 7.0 & 3.1 & 93 \\ 
12 & 0.124 & 0.352 & 9.2 & 4.8 & 71 & 0.914 & 0.956 & 6.1 & 3.1 & 98 \\ 
\midrule
\textbf{Mean} & 0.048 & 0.185 & 5.0 & 1.8 & 82 & 2.579 & 1.545 & 9.3 & 3.7 & 74 \\
\bottomrule
\end{tabular}
}  %
\end{table}

\vspace{1cm}  %

\begin{table}[h!]
\centering
\caption{Prediction errors for experiments completely unknown to the model, evaluated for both pressure and flow rate. Metrics include: MSE (mean squared error), RMSE (root mean squared error), RL2N (relative $L^2$ norm error), RL2N-B (relative $L^2$ norm error with respect to CI90 bounds), and PIB (percentage of points within bounds)}
\label{tab:error_only_val}
\resizebox{\textwidth}{!}{  %
\begin{tabular}{c|ccccc|ccccc}
\toprule
\multirow{2.5}{*}{\textbf{Experiment}} & \multicolumn{5}{c|}{\textbf{Pressure}} & \multicolumn{5}{c}{\textbf{Flow rate}} \\
                    & \textbf{MSE} & \textbf{RMSE}&\textbf{RL2N} [$\%$] &\textbf{RL2N-B} [$\%$]&\textbf{PIB [\%]}          & \textbf{MSE} & \textbf{RMSE}&\textbf{RL2N} [$\%$]&\textbf{RL2N-B} [$\%$] &\textbf{PIB [$\%$]}      \\
\midrule
1-val & 0.838  & 0.915 & 28.2 & 15.8 & 20.00 & 10.281 & 3.206 & 18.0 & 6.7 & 50.00 \\  %
2-val & 0.709  & 0.842 & 24.2 & 11.9 & 46.00 & 12.593 & 3.549 & 20.9 & 7.8 & 35.00 \\  %
3-val & 0.250  & 0.500 & 16.1 & 5.9 & 47.00 & 13.845 & 3.721 & 20.9 & 5.1 & 55.00 \\  %
4-val & 0.171  & 0.414 & 13.3 & 4.3 & 54.00 & 1.737  & 1.318 & 6.6 & 1.3 & 84.00 \\  %
5-val & 0.488  & 0.699 & 22.3 & 13.1 & 52.00 & 6.323  & 2.515 & 13.5 & 3.7  & 37.00 \\  %
6-val & 0.100  & 0.317 & 9.9 & 2.9 & 64.00 & 6.531  & 2.555 & 12.6 & 5.7 & 49.00 \\  %
7-val & 0.273  & 0.523 & 16.8 & 7.1 & 54.00 & 4.876  & 2.208 & 12.7 & 2.8 & 63.00 \\  %
8-val & 0.397  & 0.630 & 16.4 & 4.8 & 56.00 & 9.649  & 3.106 & 17.9 & 5.4 & 45.00 \\  %
\midrule
\textbf{Mean} & 0.403  & 0.605 & 18.4 & 8.2 & 49.13 & 8.229 & 2.772 & 15.4 & 4.8 & 52.25 \\ 
\bottomrule
\end{tabular}
}  %
\end{table}

\end{landscape}

\newpage

\section{Conclusion}
\label{sec:conc}
This study introduced an \ac{ML}-based \ac{DT} framework to predict key operational parameters of chamber filter presses, specifically pressure and flow rates. The framework was evaluated using two datasets: one for training and validation, and another with unknown experimental configurations.

\textbf{Summary of evaluation:}
\begin{itemize}
    \item \textbf{Pressure prediction (training \& validation):} Overall \ac{MSE} is $0.048$, \ac{RMSE} was $0.185$ and RL2N of $5.0\%$. Deviation from the CI90$\%$ bounds was $1.8\%$.
    \item \textbf{Flow rate prediction (training \& validation):} Overall \ac{MSE} is $2.579$, \ac{RMSE} was $1.545$ and RL2N of $9.3\%$. Deviation from the CI90$\%$ bounds was $3.7\%$.
    \item  \textbf{Pressure prediction (unknown data):} Overall \ac{MSE} was $0.403$, \ac{RMSE} was $0.605$ and RL2N of $18.4\%$. Deviation from the CI90$\%$ bounds was $8.2\%$.
    \item  \textbf{Flow rate prediction (unknown data):} Overall \ac{MSE} was $8.229$, \ac{RMSE} was $2.772$, and RL2N was $15.4\%$. Deviation outside CI90$\%$ bounds was $4.8\%$.
\end{itemize}

The comparison of \ac{RMSE} and \ac{MSE} values suggests fewer significant errors and more frequent small deviations in flow rate predictions. Furthermore, the model's ability to interpolate between known configurations (e.g. $12.5 g/L$ and $25 g/L$) to approximate unknown configurations (e.g. $15 g/L$) highlights its practical utility, even in cases with limited training data. However, errors for specific configurations, such as $6.25 g/L$, indicate that broader training datasets are required for better performance in edge cases. 
This \ac{DT} framework, which integrates real-time data and predictive analytics, represents a significant step forward in optimizing chamber filter press operations. By accurately forecasting pressure and flow rates and estimating the lifespan of the filter medium, the system enables more efficient process control, reduces downtime, enhances resource utilization, and supports sustainable practices. Future work will focus on expanding the diversity of operational scenarios, improving prediction accuracy for sparsely trained configurations, and refining the model’s adaptability to evolving industrial conditions.

\section*{Acronyms}
\begin{acronym}
    \acro{AI}{artificial intelligence}
    \acro{AR}{augmented reality}
    \acro{CI90}{90 Confidence Interval}
    \acro{CFD}{computational fluid dynamics}
    \acro{DT}{digital twin}
    \acro{FFNN}{feed forward neural network}
    \acro{LBM}{lattice Boltzmann method}
    \acro{LSTM}{long short-term memory}
    \acro{MA}{moving average}
    \acro{MAE}{mean absolute error}
    \acro{M}{measured}
    \acro{MSE}{mean square error}
    \acro{ML}{machine learning}
    \acro{NN}{neural network}
    \acro{NTP}{network time protocol}    
    \acro{OPC UA} {open platform communications unified architecture}
    \acro{P}{predicted}
    \acro{PIB}{points inside bounds}
    \acro{PTP}{precision time protocol} 
    \acro{ReLU}{rectified linear unit}
    \acro{RL2N}{relative \(L^2\)-norm}
    \acro{RL2N-B}{relative \(L^2\)-norm-bounds}
    \acro{RMSE}{root mean square error}
    \acro{RNN}{recurrent neural network}

    \acro{STD}{standard deviation}
\end{acronym}

\section*{Author contribution}\noindent
\textbf{D.\ Teutscher}:
Conceptualization,
Methodology,
Validation, 
Formal analysis, 
Investigation,
Software,
Data Curation, 
Writing - Original Draft;
\textbf{T.\ Weber-Carstanjen}: 
Resources; 
\textbf{S. \ Simonis}:
Validation, 
Formal analysis, 
Writing - Review \& Editing;
\textbf{M.J.\ Krause}: 
Resources, 
Funding acquisition, 
Writing - Review \& Editing.
All authors have read and approved the final version of the manuscript. 

\section*{Acknowledgements}
 
The current research is part of the Invest BW project  BW1\_0027/01 "Filt-AR: Entwicklung einer neuen Filt-AR-Filterpresse mit wissensbasierter Regelung anhand eines digitalen Zwillings und Augmented Reality-Visualisierung“.

\section*{Declaration of generative AI and AI-assisted technologies in the writing process}
During the preparation of this work, the authors used ChatGPT 4.0 in order to enhance the readability. After using this tool, the authors reviewed and edited the content as needed and take full responsibility for the content of the published article.

\bibliographystyle{elsarticle-num} 
\bibliography{cas-refs}

\end{document}